\documentclass{article}
\makeatletter
\def\input@path{{iclr2027/}{vendor/xurl/latex/}}
\makeatother
\usepackage{iclr2027_conference,times}
\iclrfinalcopy
\usepackage[T1]{fontenc}
\usepackage[utf8]{inputenc}
\usepackage{amsmath,amssymb,booktabs,array,graphicx,tabularx,flafter}
\usepackage{placeins,capt-of}
\usepackage{xurl}
\usepackage[hidelinks]{hyperref}
\newcolumntype{L}[1]{>{\raggedright\arraybackslash}p{#1}}
\newcolumntype{C}[1]{>{\centering\arraybackslash}m{#1}}
\newcommand{\tableformat}{\setlength{\tabcolsep}{4pt}\renewcommand{\arraystretch}{1.1}}

\hypersetup{
  pdftitle={The Right Source Is Not Enough: Understanding History Interference in Speech-LLM Memory},
  pdfauthor={HongJin Song, Weiqiao Shan, Runwu Shi, Yifei Wu, Yujin Wang, Jiale Luo, Jiasheng Kuang, Xinyu Yang, Zhaoqi Cui, Chuizheng Kong, Ziyu Wu, Guowu Tan, Qiuyu Fang, Chunxiang Jin},
  pdfsubject={},
  pdfkeywords={}
}

\title{\fontsize{16}{20}\selectfont
The Right Source Is Not Enough: Understanding\\
History Interference in Speech-LLM Memory}
\newcommand{\ThirdAffiliation}{Institute of Science Tokyo}
\newcommand{\affilmark}[1]{\textsuperscript{\mdseries #1}}
\author{\parbox[t]{\dimexpr\textwidth-2\tabcolsep\relax}{\raggedright
  HongJin Song\affilmark{1},\enspace Weiqiao Shan\affilmark{2},\enspace
  Runwu Shi\affilmark{3},\enspace Yifei Wu\affilmark{4},\enspace Yujin Wang\affilmark{5}\\[1pt]
  Jiale Luo\affilmark{6},\enspace Jiasheng Kuang\affilmark{1},\enspace
  Xinyu Yang\affilmark{1},\enspace Zhaoqi Cui\affilmark{1},\enspace Chuizheng Kong\affilmark{1}\\[1pt]
  Ziyu Wu\affilmark{1},\enspace Guowu Tan\affilmark{1},\enspace
  Qiuyu Fang\affilmark{1},\enspace Chunxiang Jin\affilmark{4}\\[6pt]
  \normalfont
  \begin{tabular}{@{}l@{\qquad}l@{}}
    1.~Beijing Institute of Technology & 2.~Northeastern University, China\\
    3.~\ThirdAffiliation & 4.~Ant Group\\
    5.~Wuhan University & 6.~Sichuan University
  \end{tabular}
}}
\date{}

\begin{document}
\maketitle
\fancyhead{}
\renewcommand{\headrulewidth}{0pt}
\begin{abstract}
Speech large language models retain past utterances to support multi-turn conversations. Yet remembering an event does not ensure that its information is used only when relevant. We investigate why irrelevant utterances in the dialogue history change answers even when the queried event remains represented. Using controlled changes to dialogue history, linear probes, and activation patching in MOSS-Audio and MiMo-Audio, we find that source associations remain decodable on most distractor errors. In multiple-choice QA, interference can persist in question and option states after later access to the irrelevant utterance is blocked, whereas corresponding open-ended errors nearly disappear under the tested blocking conditions. The score shifts induced by early feed-forward gate interventions depend largely on the attention Key vectors at answer option positions. Redistributing attention over the option tokens shifts answer scores even when the corresponding Values and total attention to these tokens are held fixed. On matched pairs where changing only an irrelevant utterance flips correctness, patching these Keys from a correct run repairs most errors, while jointly transferring Keys and Values from the incorrect run often reproduces the same wrong choice. These findings show that recoverable source associations do not ensure correct evidence use: irrelevant past utterances can influence answer selection through later attention over competing options.

\end{abstract}
\section{Introduction}
\label{sec:introduction}

Speech large language models (Speech-LLMs) answer questions about past utterances \citep{yang2026moss,xiaomi2025mimo}. Evaluations often test whether a model recalls an earlier utterance or sound \citep{kim2025contextdialog,xiao2026envmem}. Yet an assistant can remember a library's opening time accurately and give it as the departure time of a flight. The fact itself is correct; its relevance depends on the question.

Prior work examines multi-turn reliability, long-term memory and distracting text in comprehension, reasoning and retrieval-augmented question answering (QA) \citep{laban2026lost,wu2025longmemeval,jia2017adversarial,shi2023distracted,yoran2024irrelevant}. We trace how a remembered fact from a distractor utterance becomes evidence for an answer.

\begin{figure}[htbp]
\centering
\includegraphics[width=\linewidth]{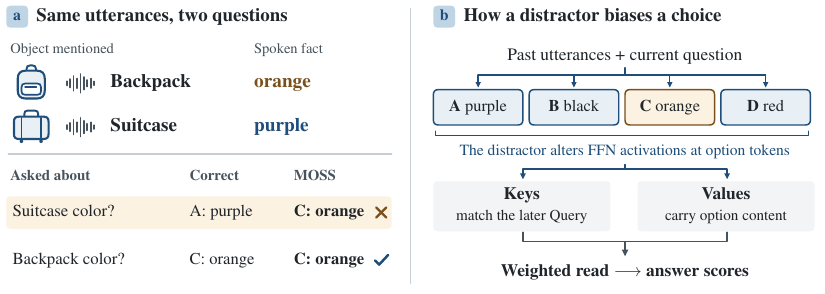}
\caption{(a) Given the same past utterances and options, MOSS selects orange for both objects, though only the backpack is orange (abridged from Table~\ref{tab:recorded-cases}). (b) A distractor alters feed-forward network (FFN) activations at option tokens. Later attention uses option Keys to weight their Values. This schematic summarizes paired interventions, not measurements on the example in (a).}
\label{fig:overview}
\end{figure}

In Figure~\ref{fig:overview}, MOSS answers \emph{orange} for both an orange backpack and a purple suitcase. A recall test about the backpack alone would miss the error. In both MOSS-Audio and MiMo-Audio, distractor utterances that name incorrect options reduce accuracy on natural-speech questions. With the number of utterances fixed, repeated mentions favor the particular wrong option named (Section~\ref{sec:question-dependent-use}).

We block later access to the distractor utterance and prevent newly generated tokens from reading it. On matched questions, the models then rarely generate the distractor answer without options, yet many original multiple-choice errors persist (Section~\ref{sec:format}). We trace how a distractor can determine a multiple-choice answer and test two implications for training.
\begin{enumerate}
\item A past utterance favors its matching option more when it is relevant. On most distractor errors, probes still identify the relevant utterance and match each utterance to an option. The model retains the information needed to distinguish the sources.
\item Early FFN gates can change the Keys that later attention uses to weight option Values. When changing only the distractor flips a correct answer, borrowing the correct run's option Keys repairs most such errors without removing either utterance. Transferring the wrong run's option Keys and Values together often induces the same wrong choice (Section~\ref{sec:mechanism}).
\item When a fact changes, train on two questions: one answer should change, the other should not. This reduced distractor choices on new scenarios using the same templates. Adjusting each audio clip's Value contribution also improved accuracy on MMAR-derived questions with multiple clips. Neither test used a source cue (Section~\ref{sec:implications}).
\end{enumerate}

\section{Experimental Setup}
\label{sec:framework}

\paragraph{Data and comparisons.}
Each controlled dialogue history contains two synthetic utterances. We change a detail in one and keep the other, the question and all four options fixed (Table~\ref{tab:paired-example}).

\begin{table}[htbp]
\centering
\caption{The backpack pair in the four-version design (Appendix~\ref{app:paired-materials}). Each question is asked separately.}
\label{tab:paired-example}
\tableformat
\begin{tabularx}{\linewidth}{@{}L{.22\linewidth}>{\raggedright\arraybackslash}X@{}}
\toprule
\textbf{Input} & \textbf{Content} \\
\midrule
Fixed utterance & Amelia Campbell's suitcase has a green color. \\
Changed utterance & Amelia Campbell's backpack has a white color.\newline Paired version: Amelia Campbell's backpack has a yellow color. \\
\midrule
Suitcase question & What color is Amelia Campbell's suitcase? \\
Backpack question & What color is Amelia Campbell's backpack? \\
\midrule
Answer options & A: green; B: blue; C: white; D: yellow. \\
\bottomrule
\end{tabularx}
\end{table}

The changed utterance describes one of two events or objects, each with two possible answer contents. These four versions, both questions and both utterance orders give 8,192 inputs per model across 512 held-out scenarios. Option order is fixed within pairs \citep{liu2024lostmiddle,pezeshkpour2024order}; intervals resample groups that share utterances.

We group LibriSpeech utterances \citep{panayotov2015librispeech} into 1,000 sets of three. For each clip, we ask which of three words it contains (3,000 questions). The other two utterances each name a different incorrect option in the competing condition; neither names an option in the neutral condition.

The relevant utterance stays fixed. Only what the model hears in the other utterances changes.

\paragraph{Models and measurements.}
MOSS-Audio-8B-Instruct passes encoder outputs through an adapter \citep{yang2026moss}; MiMo-Audio-7B-Instruct encodes each group of audio tokens as one input vector \citep{xiaomi2025mimo}. We score choices by the highest next-token option logit and report accuracy and distractor selections. Without options, we score the generated answer against the relevant and distractor answers, allowing a match to neither.

We examine the feed-forward network (FFN) gate activation $g$ \citep{shazeer2020glu} at each option's label, text and trailing separator. Query, Key and Value denote attention vectors; an answer value is a detail such as a time or color. Both decoders have 36 layers. We fix the boundary after layer 17 so that interventions on early option computation and later attention reads do not overlap. This split was not selected on test errors and does not identify exclusive functional stages (Appendix~\ref{app:layer-choice}).

We transfer activations between matching positions in both directions \citep{meng2022locating,zhang2024patching}. In Table~\ref{tab:paired-example}, the white-backpack run (the receiver) keeps its own audio and takes an activation from the yellow-backpack run (the donor). Let $\ell_y$ and $\ell_x$ be the logits for yellow (D) and white (C). We measure
\begin{equation}
\Delta s = \bigl[\ell_y-\ell_x\bigr]_{\text{after patching}}
          -\bigl[\ell_y-\ell_x\bigr]_{\text{before patching}}.
\label{eq:directed-score}
\end{equation}
Here, $\Delta s>0$ means yellow gained on white.

Yellow and white can exchange ranks while green still wins. We therefore also count flips from a correct answer to the donor's exact wrong option.

\section{Characterizing History Interference}
\label{sec:behavior}

\subsection{Distractor-Induced Answer Bias}
\label{sec:question-dependent-use}

In the MOSS example in Table~\ref{tab:recorded-cases}, one utterance says the backpack is orange and the other says the suitcase is purple. MOSS correctly chooses \emph{orange} for the backpack but also chooses \emph{orange} for the suitcase. The model gives a color it heard, but that color belongs to the backpack.

\begin{table}[!htbp]
\centering
\caption{Two questions per dialogue history. Only the object in the question changes; the utterances and options stay fixed. The last two columns show the correct option and the model's choice.}
\label{tab:recorded-cases}
\tableformat
\begin{tabularx}{\linewidth}{@{}L{.075\linewidth}>{\raggedright\arraybackslash}X@{\hspace{12pt}}C{.115\linewidth}C{.115\linewidth}@{}}
\toprule
\textbf{Model} & \textbf{Utterances, options and questions} & \textbf{Correct} & \textbf{Selected}\\
\midrule
MOSS &
\textit{Utterance 1:} When asked about the backpack, Camille Russell said its color is orange.\newline
\textit{Utterance 2:} When asked about the suitcase, Camille Russell said its color is purple.\newline
\textit{Options:} A: purple; B: black; C: orange; D: red.
& & \\
\cmidrule(l){2-4}
& What color is Camille Russell's suitcase? & A: purple & C: orange\\
\cmidrule(l){2-4}
& What color is Camille Russell's backpack? & C: orange & C: orange\\
\midrule
MiMo &
\textit{Utterance 1:} To keep it safe, Abigail Thomas put the boarding pass inside the locker.\newline
\textit{Utterance 2:} To keep it safe, Abigail Thomas put the passport inside the box.\newline
\textit{Options:} A: box; B: locker; C: suitcase; D: drawer.
& & \\
\cmidrule(l){2-4}
& Where does Abigail Thomas keep the passport? & A: box & B: locker\\
\cmidrule(l){2-4}
& Where does Abigail Thomas keep the boarding pass? & B: locker & B: locker\\
\bottomrule
\end{tabularx}
\end{table}

\begin{figure}[!htb]
\centering
\includegraphics[width=\linewidth]{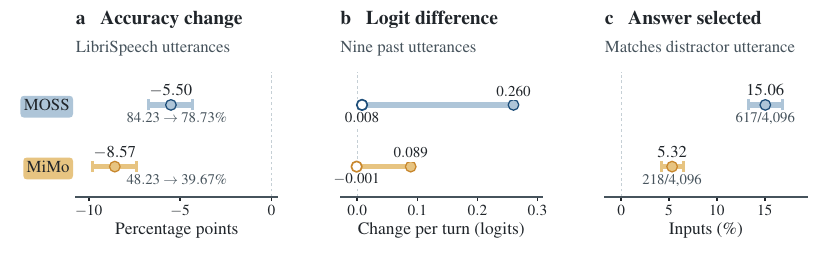}
\caption{Effects of distractor utterances on answers. (a) Change in accuracy on 3,000 natural-speech questions when distractor utterances that mention no options are replaced with utterances that name incorrect options. (b) Mean change in the specified incorrect option's logit minus the correct option's logit per additional utterance that mentions the specified option (filled circles) or another incorrect option (open circles). (c) Percentage of 4,096 answers per model that match the option named in the distractor utterance. Error bars show 95\% confidence intervals, resampling whole dialogue histories in (a) and groups that share audio in (c). Panel (b) shows point estimates. See Tables~\ref{tab:behavior-estimates} and~\ref{tab:relevant-behavior}.}
\label{fig:behavior-evidence}
\end{figure}

Across natural-speech questions, replacing two neutral utterances with two that each name a different incorrect option lowers accuracy by 5.50 percentage points in MOSS and 8.57 in MiMo (Figure~\ref{fig:behavior-evidence}a). The question and relevant utterance are unchanged. Accuracy also falls on the subset that both models answer correctly from the relevant utterance alone (Table~\ref{tab:behavior-estimates}). With reference transcripts, accuracy across all questions drops by 1.53 percentage points in MOSS and 17.93 in MiMo (Appendix~\ref{app:librispeech-interference}). The observed drop is larger with audio for MOSS but with text for MiMo.

Holding the total at nine utterances, we replace neutral utterances with ones that name an incorrect option. Each additional mention of $x$ raises its score relative to the correct option on average; a mention of another incorrect option $y$ leaves that same margin almost unchanged (Figure~\ref{fig:behavior-evidence}b).

In Table~\ref{tab:paired-example}, changing the backpack from white to yellow should change the backpack answer but leave the suitcase answer green. Across 2,048 pairs asking about the changed detail, both answers are correct in 76.51\% of MOSS pairs and 93.31\% of MiMo pairs. When asked about the unchanged object or event, the models instead select the option named in the distractor on 15.06\% and 5.32\% of 4,096 inputs, respectively (Figure~\ref{fig:behavior-evidence}c; Table~\ref{tab:relevant-behavior}).

\FloatBarrier

\noindent The model can treat a fact about the wrong object as evidence. The backpack's color says nothing about the suitcase's. Lowering only the correct score would favor all wrong options equally. A match to an irrelevant utterance can instead give a particular option an advantage.

\subsection{Interference Across Answer Formats}
\label{sec:format}

We ask the same question about the same utterances twice. Without options, the model generates its answer; with four options, we take the highest-scoring label. Both models sometimes give the answer from the distractor utterance without options and do so more often with options (Table~\ref{tab:access-states}).

\begin{table}[!htbp]
\centering
\caption{Number of incorrect answers that match the answer given in the distractor utterance. The distractor is always the first utterance; the mask covers its audio tokens and surrounding special tokens. \emph{All inputs} counts every such error. The last two columns count how many original errors recur in each condition. Appendix Figure~\ref{fig:state-bridge} shows which hidden states are copied.}
\label{tab:access-states}
\tableformat
\begin{tabularx}{\linewidth}{@{}>{\raggedright\arraybackslash}X *{6}{C{.093\linewidth}}@{}}
\toprule
& \multicolumn{2}{c}{\textbf{Open-ended}} & \multicolumn{4}{c}{\textbf{Multiple choice}}\\
\cmidrule(lr){2-3}\cmidrule(l){4-7}
& \multicolumn{2}{c}{All inputs} & \multicolumn{2}{c}{All inputs} & \multicolumn{2}{c}{\shortstack{Originally selected the\\other utterance's answer}}\\
\cmidrule(lr){2-3}\cmidrule(lr){4-5}\cmidrule(l){6-7}
\textbf{Condition} & MOSS & MiMo & MOSS & MiMo & MOSS & MiMo\\
\midrule
No intervention & 151 & 22 & 175 & 94 & 175 & 94\\
Mask distractor attention & 2 & 3 & 131 & 57 & 124 & 49\\
\midrule
Copy states: early attention blocked & \textemdash & \textemdash & 6 & 2 & 6 & 1\\
Copy states: early attention allowed & \textemdash & \textemdash & 132 & 58 & 125 & 49\\
\bottomrule
\end{tabularx}
\end{table}

We test whether these errors persist without later access to the distractor. Prompt tokens after the utterance can read it in layers 0--17 but not 18--35 \citep{geva2023dissecting} (Appendix~\ref{app:layer-choice}). In open-ended QA, newly generated tokens cannot read it at any layer. Among 2,048 matched inputs, MOSS gives the distractor answer in 2 open-ended responses versus 131 multiple-choice responses; MiMo does so in 3 versus 57. The difference persists when a single token suffices for either answer: among the original distractor errors on the 512 color inputs, none of 32 open-ended errors recur after blocking, compared with 37 of 51 multiple-choice errors in MOSS.

We then fix the options and block later access in two runs of each multiple-choice question. Only one run lets the question and option tokens read the distractor in the early layers. After layer 17, we copy hidden states between corresponding text positions, from the current user-role marker through the next-to-last input token \citep{zhang2024patching}. This excludes the utterances and the final answer-scoring position. With states from the run that blocked early access, only 6 of 175 original distractor errors recur in MOSS and 1 of 94 in MiMo. Reversing the transfer brings back 125 and 49, respectively.

\FloatBarrier

\noindent Under the tested blocking conditions, interference is more persistent when options are listed. Early exposure rarely suffices to generate the distractor answer in open-ended QA. In multiple choice, that answer is already in the prompt. If choosing it required rereading the utterance, restoring early text states could not bring back the same wrong answer with the options fixed and later access blocked. The states formed while processing the question and options can carry a preference for the distractor answer into the final choice.

\section{Mechanisms of History Interference}
\label{sec:mechanism}

\subsection{Probing Source Associations}
\label{sec:retained}

\begin{figure}[!htbp]
\centering
\includegraphics[width=\linewidth]{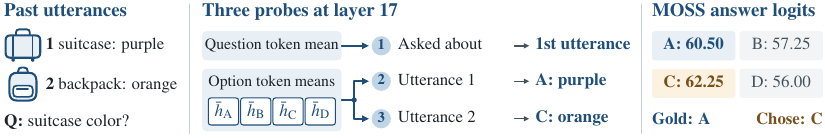}
\caption{Probe predictions and MOSS's own answer after swapping the utterances in Table~\ref{tab:recorded-cases}.}
\label{fig:probe-case}
\end{figure}

Asked about a purple suitcase, MOSS chooses orange, the backpack color (Figure~\ref{fig:probe-case}). We read the associations at layer 17. One linear probe identifies the relevant utterance from the question tokens. Two others use the option tokens to identify which option matches each utterance. All are fitted on development data and frozen; the model answers independently.

All three probes are correct on 488/617 MOSS and 206/218 MiMo distractor errors when the question asks about the unchanged utterance (Table~\ref{tab:source-recovery-main}). In most of the 4,096 pairs, all three are correct in both orders: the answer words stay the same, but their source assignments reverse.

\begin{table}[!htbp]
\caption{Probe accuracy (\%). The error column includes only questions about the unchanged utterance answered with the distractor option. Random labels test the question probe. Intervals: Appendix~\ref{app:readout-results}.}
\label{tab:source-recovery-main}
\begin{center}\small\tableformat
\begin{tabular*}{\linewidth}{@{\extracolsep{\fill}}lcccc@{}}
\toprule
Model & \shortstack{All three on\\distractor errors} & Question (8,192 inputs) & Random (16 fits) & All three, both orders\\
\midrule
MOSS & 79.09 & 93.73 & 49.98 & 85.23\\
MiMo & 94.50 & 99.58 & 49.95 & 97.46\\
\bottomrule
\end{tabular*}
\end{center}
\end{table}

\FloatBarrier

\noindent On errors where all three probes are correct, choosing the option matched to the utterance identified by the question probe would give the right answer. The states thus support a decision rule that the decoder need not use \citep{ravichander2021probing}. When only the correct option was mentioned, remembering what was heard and answering from the relevant utterance give the same answer. When both options were mentioned, the question determines which mention counts as evidence. Accuracy alone misses this distinction: if changing only the distractor utterance narrows the correct option's lead, irrelevant content has influenced the scores even when the winner stays the same.

Complementary subspace tests add a content-contrast edit to a question-contrast edit. This favors the unasked answer on average at all five prespecified layers under both fitting orders (Table~\ref{tab:conditional-content-order}). Order can change directions and donor coefficients (Appendix~\ref{app:subspace-order}, Eq.~\ref{eq:subspace-order-difference}); our probes and subsequent component tests do not use these fitted subspaces.

\subsection{FFN Gating and Option Keys}
\label{sec:candidate-formation}

A wrong choice need not mean that the correct option lost support. In Table~\ref{tab:paired-example}, the two utterances describe a green suitcase and a white backpack; the question asks about the suitcase. At option tokens in layers 0--17, we replace only the FFN gate $g$ with $g'$ from the yellow-backpack run. The change $W_d((g'-g)\odot u)$ reweights the current run's features $u$. We test both directions.

Gate transfer shifts the scores toward the paired distractor (yellow over white), while the correct option barely moves relative to the unmentioned option (green versus blue; Table~\ref{tab:option-margin-patch}). Permuting the gate changes across channels preserves their size but largely removes the preference: its direction depends on which features are reweighted. Appendix~\ref{app:ffn-position-controls} compares option and question positions.

\begin{table}[htbp]
\centering\small\tableformat
\caption{Mean logit changes after gate transfer. The first three contrasts use the unmentioned option as reference.}
\label{tab:option-margin-patch}
\begin{tabular*}{\linewidth}{@{}l@{\extracolsep{\fill}}rrrr@{}}
\toprule
Model & Receiver distractor & Donor distractor & Correct & Donor minus correct \\
\midrule
MOSS & $-1.771$ & $+1.966$ & $+0.038$ & $+1.93$ \\
MiMo & $-0.781$ & $+0.817$ & $-0.002$ & $+0.82$ \\
\bottomrule
\end{tabular*}
\end{table}

\begin{figure}[!htb]
\centering
\includegraphics[width=\linewidth]{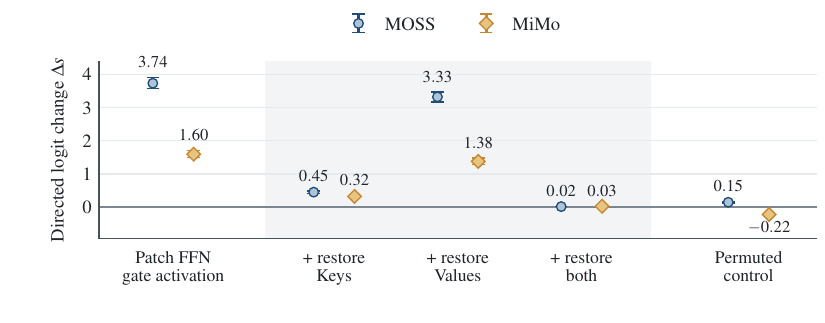}
\caption{Mean change in the donor-minus-receiver distractor logit difference $\Delta s$ over 4,096 inputs per model (yellow minus white in the example); bars: 95\% audio-group intervals. Fraction of the gate-only shift retained (MOSS/MiMo): Keys restored, 12\%/20\%; Values restored, 89\%/86\%.}
\label{fig:gate-reading}
\end{figure}

To separate reading weights from content, we keep the early gates patched. When tokens after the options read them in layers 18--35, we restore either option Keys or option Values from the unedited run. Restoring Keys removes most of the bias; restoring Values leaves most of it intact (Figure~\ref{fig:gate-reading}). Neither intervention blocks access to the utterances.

\begin{samepage}
\noindent On average, transferring the early gates raises the relative score of the paired distractor (yellow) but barely changes that of the correct option (green). The gate weights features, not individual utterances: if two sources contribute to one feature, its coefficient cannot scale their contributions separately. If gating makes the Key at a yellow-option token a better match for a later Query relative to other Keys, that token's Value receives more weight. The model can retain green as the suitcase color yet count the backpack color as evidence about the suitcase.
\par
\end{samepage}

\subsection{Option Attention Allocation}
\label{sec:candidate-reading}

\begin{figure}[!htb]
\centering
\includegraphics[width=\linewidth]{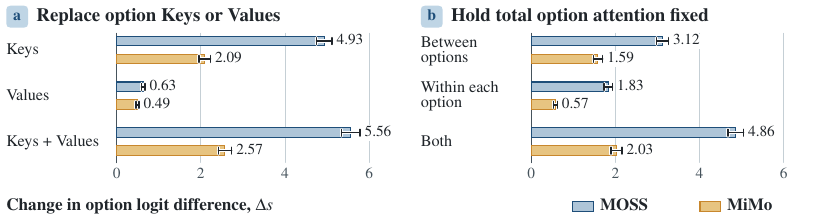}
\caption{Option reads without a gate replacement. Bars show mean $\Delta s$ over 4,096 distractor-change inputs per model; whiskers show 95\% audio-group intervals. A positive $\Delta s$ favors the paired distractor.}
\label{fig:reading-components}
\end{figure}

Borrowing gates from a correct run repairs 251/273 MOSS and 104/123 MiMo paired errors, but changes the option states themselves. To isolate the read, we keep the white-backpack input and its option states intact. In layers 18--35, only tokens after the options read option Keys from the yellow-backpack run. Separate runs substitute Values or both. Each reading token computes its own Query. Across pairs, transferring Keys shifts scores toward the paired distractor more than transferring Values (Figure~\ref{fig:reading-components}a).

For each head and current Query we also fix total option weight, non-option weights and Values. With the paired Keys we change shares \emph{between} options, token proportions \emph{within} each option, or both (Appendix~\ref{app:option-read-method}). Between-option changes have the larger mean effect. Changing both nearly matches the shift from transferring option Keys: 4.86 versus 4.93 logits in MOSS and 2.03 versus 2.09 in MiMo (Figure~\ref{fig:reading-components}b).

The score shift toward the paired fact is larger when the question asks about that fact, but remains positive on average when the question asks about the other utterance (Table~\ref{tab:asked-unasked-path}).

\begin{table}[!htb]
\centering
\caption{Mean $\Delta s$ when the question concerns the changed utterance / the other utterance (4,096 inputs each). The interventions are run separately.}
\label{tab:asked-unasked-path}
\tableformat
\begin{tabular*}{\linewidth}{@{}l@{\extracolsep{\fill}}rrrr@{}}
\toprule
Model & Early FFN gate & Option Keys & Option Values & Keys and Values\\
\midrule
MOSS & 7.373 / 3.737 & 9.520 / 4.933 & 2.076 / 0.632 & 11.621 / 5.564\\
MiMo & 6.623 / 1.598 & 9.633 / 2.093 & 2.189 / 0.492 & 11.864 / 2.573\\
\bottomrule
\end{tabular*}
\end{table}

\noindent More attention to yellow need not raise its score. At fixed Values and within-option proportions, shifting weight $\delta$ from white to yellow changes one head's output by $\delta(\mu_{\mathrm{yellow}}-\mu_{\mathrm{white}})$, where $\mu_j$ is option $j$'s weighted mean Value \citep{kobayashi2020weight}. Shared components cancel; only differences the decoder uses to distinguish these answers affect their score gap. Favoring the new backpack color serves the backpack question. On suitcase questions, the weaker shift raises yellow relative to white although both are wrong. Even without extra attention to the utterance about the backpack, its content can still bias the answer when the model reads the options.

\subsection{Correcting and Reproducing Errors}
\label{sec:natural-errors}

\begin{figure}[!htb]
\centering
\includegraphics[width=\linewidth]{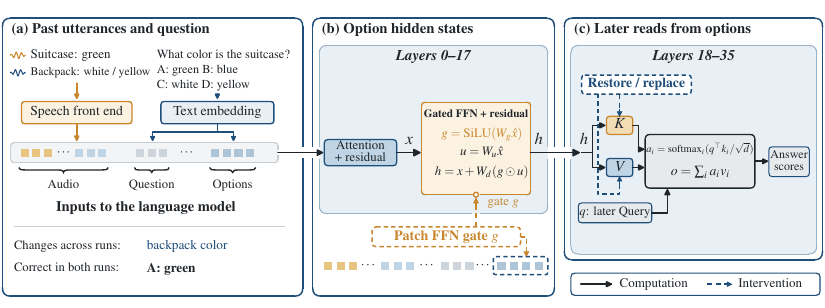}
\caption{(a) Only the distractor utterance changes. (b) Early FFN gate ($\hat x$: normalized input). (c) The Query and option Keys determine the weights on the option Values.}
\label{fig:candidate-path}
\end{figure}

To connect score shifts to actual errors, we select pairs where changing only the distractor utterance switches the answer from correct ($c$) to that utterance's wrong option ($d$). Both utterances remain accessible (Figure~\ref{fig:candidate-path}).

Borrowing the correct partner's early FFN gates or, separately, its later option Keys repairs most paired errors (Table~\ref{tab:specific-errors}, left). But $d$'s mean lead over $c$ is smaller than $c$'s lead over $d$ in the correct partner: a small shift can repair an error without explaining why that distractor won.

\begin{table}[!htb]
\centering\small\tableformat
\caption{Choices on 273 MOSS and 123 MiMo pairs. Indented rows keep the gate transfer and restore the receiving run's unedited option Keys, Values, or both.}
\label{tab:specific-errors}
\begin{tabular*}{\linewidth}{@{}l@{\extracolsep{\fill}}rrrr@{}}
\toprule
& \multicolumn{2}{c}{Corrected ($d\to c$)} & \multicolumn{2}{c}{Reproduced ($c\to d$)}\\
\cmidrule(lr){2-3}\cmidrule(l){4-5}
Intervention & MOSS & MiMo & MOSS & MiMo\\
\midrule
None & 0 & 0 & 0 & 0\\
Early FFN gate & 251 & 104 & 51 & 11\\
\quad + restore option Keys & 99 & 40 & 0 & 0\\
\quad + restore option Values & 253 & 103 & 23 & 9\\
\quad + restore option Keys and Values & 16 & 11 & 0 & 0\\
Option Keys & 269 & 113 & 110 & 38\\
Option Values & 146 & 67 & 0 & 1\\
Option Keys and Values & 256 & 113 & 240 & 87\\
\bottomrule
\end{tabular*}
\end{table}

In reverse, only a switch to the wrong partner's \emph{specific} $d$ counts (Table~\ref{tab:specific-errors}, right). Its gates seldom induce $d$ alone. Restoring the correct run's option Keys prevents those switches. Transferring the wrong run's Keys and Values together produces $d$ in most pairs, even among those with decodable source associations in both runs (Section~\ref{sec:retained}).

Joint transfer changes both the weights and the content they carry. Compared with gate transfer, it raises $d$ more than it lowers $c$, relative to the unmentioned option (Table~\ref{tab:paired-error-details}). When joint Key/Value transfer reproduces the error but neither single transfer does, adding the separate shifts to each option's original logit predicts the same distractor answer in 61/133 MOSS pairs and 35/48 MiMo pairs (Appendix~\ref{app:interactions}). Jointly crossing the decision boundary therefore does not, by itself, establish a nonlinear interaction.

\noindent At option tokens in layers 0--17, the correct partner's FFN gates reweight the wrong run's features as they enter the option states. Most gate repairs disappear when tokens after the options use the wrong run's original Keys in layers 18--35, but survive when its Values are restored instead. Much of the early correction thus depends on how later tokens read the options. Leaving the option states intact and borrowing only correct-run Keys lets each Query reweight the same Values and repairs most paired errors. Borrowing Values instead changes the content passed on under the receiver's Keys and repairs fewer. Even after the backpack color has influenced the options, changing the later read can correct the suitcase answer without removing either utterance.

\FloatBarrier

\subsection{Learning Selective History Use}
\label{sec:implications}

Borrowing option Keys can correct an error while both utterances remain available (Sections~\ref{sec:retained}--\ref{sec:natural-errors}). Without a correct partner, the question must guide which reads to change.

Let $m=\ell_d-\ell_c$ be the distractor-minus-correct score, $o=\sum_i a_i v_i$ one head's weighted sum of Values \citep{vaswani2017attention}, and $r=\nabla_o m$ the sensitivity of the final score difference to that sum. At fixed Query $q$, changing the Keys and Values at option positions $i\in C$ by $\epsilon\,\delta k_i$ and $\epsilon\,\delta v_i$ gives (Appendix~\ref{app:attention-derivative}):
\begin{equation}
\left.\frac{d m}{d\epsilon}\right|_0
=\underbrace{\sum_{i\in C}a_i\,r^\top\delta v_i}_{\text{changed Values}}
+\underbrace{\sum_{i\in C}a_i\,r^\top(v_i-o)
\frac{q^\top\delta k_i}{\sqrt{d_k}}}_{\text{changed weights}}.
\label{eq:content-weighting}
\end{equation}
Lowering one option token's attention score redistributes weight to other positions. With other attention scores and Values fixed, lowering this score reduces $m$ to first order exactly when $r^\top(v_i-o)>0$: that Value must favor the distractor more than the weighted average does. The right adjustment therefore depends on what the token contributes to this answer, not just on its weight. In the paired experiments, transferring option Keys or Values shifts scores toward the content of the donor utterance more when the question asks about that content (Table~\ref{tab:asked-unasked-path}).

\paragraph{Training on which answers should change.}
For each scenario, we change a fact in one utterance and ask the same two questions about each version, keeping the options fixed and giving no source cue \citep{gardner2020contrast,kim2025contextdialog}. For example, when the backpack color changes, the backpack answer must change and the suitcase answer must stay green. We train rank-8 LoRA adapters \citep{hu2022lora} in the decoder's attention and FFN projections using ordinary cross-entropy over the four options. On new scenarios with the same templates and voices, both models more often answer all four inputs correctly and less often choose the distractor (Table~\ref{tab:uncued-source-use}a; Appendix~\ref{app:source-use-validation}).

\paragraph{Adjusting each audio clip's contribution.}
In a separate experiment, we freeze the backbone and add a rank-32 adapter to eight attention heads per layer in layers 9--17. While processing the question turn, each selected head computes the contribution of audio turn $s$ as $o_s=\sum_{i\in s}a_i v_i$, using its ordinary attention weights. From the current Query, $o_s$, and the sum across audio turns, the adapter computes a correction for each turn and adds the corrections to the head output. This leaves the attention weights for that read unchanged. We train the adapter on answer labels, with a paired loss that favors agreement when the correct answer stays fixed and a shift toward the new answer when it changes.

Without a source cue, the adapter improves both models' accuracy on MMAR-derived questions with multiple audio clips (\citealp{ma2025mmar}; Table~\ref{tab:uncued-source-use}b). Accuracy also rises with the relevant audio clip alone, so the overall gain cannot be attributed specifically to suppressing interference (Appendix~\ref{app:value-read-adaptation}). Neither training experiment evaluated transcription or open-ended dialogue after adaptation.

\begin{table}[!htb]
\centering\small\tableformat
\caption{Selective use without a source cue. (a) Controlled QA (\%): 8,192 answers in 2,048 sets; supervised means over three seeds. (b) 517 MMAR-derived questions with two to four audio clips; 95\% intervals paired by parent video.}
\label{tab:uncued-source-use}
\begin{tabular*}{\linewidth}{@{}l@{\extracolsep{\fill}}rrrrr@{}}
\toprule
(a) Model & \shortstack{Item\\accuracy} & \shortstack{Correct\\update} & \shortstack{Correct\\stability} & \shortstack{All four\\correct} & \shortstack{Distractor\\answer} \\
\midrule
MOSS & 84.12 & 76.17 & 77.98 & 60.11 & 15.72 \\
MOSS, supervised & 96.46 & 94.92 & 95.02 & 89.94 & 3.30 \\
MiMo & 95.31 & 93.26 & 91.50 & 85.30 & 4.36 \\
MiMo, supervised & 99.76 & 99.54 & 99.67 & 99.22 & 0.02 \\
\bottomrule
\end{tabular*}
\par\smallskip
\begin{tabular*}{\linewidth}{@{}l@{\extracolsep{\fill}}ccc@{}}
\toprule
(b) Model & Original (\%) & Adapted (\%) & Change (pp; 95\% CI) \\
\midrule
MOSS & 54.48 & 59.12 & $+4.64\;[1.80,7.42]$ \\
MiMo & 53.32 & 59.45 & $+6.13\;[3.20,9.16]$ \\
\bottomrule
\end{tabular*}
\end{table}

\noindent Repeating the new backpack color suffices for the backpack question. Keeping the suitcase answer unchanged tests whether the model uses the new color selectively. The adapter addresses the same distinction by adjusting each clip's contribution to the current answer.

\noindent The useful training target is how a retained utterance counts toward the current choice, not whether it remains in memory.

\section{Related Work}
\label{sec:related}

\paragraph{Remembering a source and using it.}
ContextDialog tests utterance recall and use \citep{kim2025contextdialog}; LongMemEval distinguishes memory skills in text dialogue \citep{wu2025longmemeval}. Early mistakes can persist across turns \citep{laban2026lost}. EnvMem links failed acoustic recall to representation drift \citep{xiao2026envmem}; \citet{lee2026awareness} distinguish context awareness from adherence in spoken dialogue. We ask why a fact about the wrong event can favor an option when source associations are still decodable.

\paragraph{Distractors and option evaluation.}
Distracting context changes answers in text comprehension, reasoning, and retrieval-augmented QA \citep{jia2017adversarial,shi2023distracted,yoran2024irrelevant}. \citet{cheng2025stochastic} contrast question- and context-driven computations. Other studies identify entity--attribute representations \citep{feng2024binding}, mixed retrieval strategies \citep{gurarieh2026mixing} and retrieval heads \citep{wu2025retrieval}. In multiple choice, \citet{wiegreffe2025answer} trace answer-symbol formation; \citet{wong2026decide} separate option evaluation from symbol binding.

With option order fixed within pairs and utterance order counterbalanced \citep{liu2024lostmiddle,pezeshkpour2024order}, we trace how changing what an unasked utterance says shifts the scores of the same options.

\paragraph{Tracing and controlling historical influence.}
Activation patching tests internal contributions \citep{meng2022locating,zhang2024patching}; FFNs carry predictive content \citep{geva2021ffn}; probe controls distinguish available information from classifier behavior \citep{hewitt2019probes}. H2O and SnapKV select cache entries for future reads \citep{zhang2023h2o,li2024snapkv}. We trace what persists in question and option states after direct access to the distractor is blocked, then test how later attention to the options affects the choice.

\section{Conclusion}
\label{sec:conclusion}

On most multiple-choice distractor errors in MOSS-Audio and MiMo-Audio, probes identify the relevant utterance and match each utterance to an option. Yet a fact about the wrong event can still favor the option it matches. Under the tested blocking conditions, question and option states can carry this preference after direct access to the distractor utterance ends. Patching early FFN gates shifts scores toward the distractor. The gate-induced bias passes mainly through the option Keys, which later Queries use to weight the option Values. On matched correct/error pairs, correct-run Keys repair most errors; jointly transferring wrong-run Keys and Values often reproduces the same wrong choice.

Training on which answers should change or stay fixed reduced distractor choices on new scenarios with the same templates. Adjusting per-clip Value contributions improved accuracy on MMAR-derived questions with multiple clips.

\label{last-main-page}

\clearpage
\subsection*{AI use statement}
We used AI models to search the literature, develop hypotheses, design experiments, prepare synthetic materials, and write and check analysis code. We also used them to discuss what the results imply, revise and translate the manuscript, format references, and prepare the \LaTeX{} source. All reported measurements come from experimental runs. The authors take responsibility for the analyses and the final manuscript.

\subsection*{Reproducibility statement}
Section~\ref{sec:framework} defines the materials, evaluation units, models, and measurements. Sections~\ref{sec:behavior} and~\ref{sec:mechanism} describe the comparisons and interventions. Appendix~\ref{app:connected-methods} specifies readouts, component transfers, recorded runtime, and statistical procedures; Appendix~\ref{app:component-results} gives complete component comparisons. Appendices~\ref{app:behavior-methods} and~\ref{app:subspace-controls} provide behavioral details and subspace controls, including paraphrase templates. Reproducing the full study also requires the experimental code, audio materials, and per-example outputs. After anonymous review, we will open-source all experimental scripts and release all audio materials we can redistribute, including all utterances we generated. For third-party clips that we cannot redistribute, we will provide source identifiers and scripts to retrieve and prepare them.

\subsection*{Ethics statement}
The experiments analyze released models, public speech corpora, and controlled synthetic utterances. Names in the synthetic examples are fictional experimental labels, not participant identities. We do not infer sensitive speaker attributes. The results concern how reliably models use evidence from the dialogue history; they should not be interpreted as estimates of privacy leakage or failure prevalence in unrestricted real conversations.

\bibliographystyle{iclr2027/iclr2027_conference}
\bibliography{references-draft}

\clearpage
\appendix
\section{Methods for the Option-Level Interventions}
\label{app:connected-methods}

\subsection{Paired Materials and Receiving Positions}
\label{app:paired-materials}

Each input contains a fixed utterance and one of four alternatives. We cross two events or objects (one more closely related to the fixed event or object than the other) with two possible answer contents. In Table~\ref{tab:paired-example}, the suitcase stays green. The alternatives describe a white backpack, a yellow backpack, a white garden gate, or a yellow garden gate. A content pair changes the color of the same object. We test both utterance orders and both questions for each version, giving 16 inputs and eight undirected content pairs per scenario. The test set has 4,096 undirected pairs and 8,192 directed transfers per intervention condition. The changed-distractor analysis uses the 4,096 directed inputs whose question asks about the unchanged utterance.

We used Piper 1.3.0 to synthesize all five utterance files in each scenario with the same voice and length scale. Across scenarios we alternated between \texttt{en\_US-lessac-medium} and \texttt{en\_US-ryan-medium}. The length scale cycled from 0.96 to 1.04 in steps of 0.02; both synthesis noise parameters were zero. We resampled the clips to 24~kHz, padded them with silence to the same duration within each scenario, and saved them as 16-bit PCM.

Transfers require identical question and option token sequences, equal decoder sequence lengths, and aligned source spans within a pair. A failed alignment stops execution rather than excluding the pair. No tested pair was dropped. Option positions are the union of four non-overlapping option spans, including their following separators. Question positions precede them. The option-reading interventions apply only to Query positions strictly after the final option token, including the final prompt position used to score the answer. They do not edit the audio, question, or option-side computation.

Aligned positions hold the wording and location of each option fixed. A transfer at a post-option read can change what a later token receives from an option without replacing what the model computed at that option.

\subsection{Why These Layers Were Used}
\label{app:layer-choice}

Layer indices start at zero. Both decoders have 36 layers, and the earlier probe protocol saved post-FFN states every six layers (5, 11, 17, 23, 29, 35). The state after layer 17 is both a recorded state and the midpoint of the decoder. The early-access protocol already fixed this boundary. Reusing it lets us intervene on option computation in layers 0--17, read out the state at its exit, and then intervene on later reads in layers 18--35 without overlapping the first intervention. Using the same split in both models avoids comparing different depths. Each intervention acts jointly across its specified range. Neither the split nor the individual layers were optimized on the test errors, and these comparisons do not locate an onset, a peak, or an exclusive functional stage.

The subspace and training studies use different layer choices for different questions. The subspace order audit covers all 36 layers; the paraphrase controls repeat five previously specified layers (16, 20, 22, 23 and 24). Layer 23 illustrates an order comparison, not a universal peak (Appendix~\ref{app:subspace-controls}). The separate Value adapter retains the reference training configuration in layers 9--17 for the cross-model comparison, with heads chosen per model on development utterances (Appendix~\ref{app:value-read-adaptation}). Its accuracy does not establish that this is the optimal training range.

The midpoint gives the two interventions an unambiguous order: an early change can affect a later read, but a later edit cannot change the already computed early state. This identifies a direction of influence, not a unique layer at which interference begins.

\subsection{Linear Probes}
\label{app:readout-results}

We save post-FFN states after layers 5, 11, 17, 23, 29, and 35. Question features are the mean over the question tokens. Option features concatenate the four option-wise means in the displayed order. The reported conjunction uses the requested-source classifier at question positions and two source-to-option classifiers at option positions, all at layer 17. Source labels describe which utterance is asked about; source-to-option labels describe which of the four options expresses each utterance's content. They are separate prediction tasks.

Each classifier is a full-dimensional linear ridge readout fitted to one-hot labels. Four-fold validation splits the 128 development audio groups. Centering and scaling are fitted separately inside each training fold. Regularization is selected from $10^{-6},10^{-5},10^{-4},10^{-3},10^{-2},10^{-1}$ by development accuracy, with ties favoring stronger regularization. The chosen classifier is refitted on all 2,048 development inputs and applied unchanged to the test states. No dimensionality reduction or test-based feature selection is used. Following the use of control tasks to interpret probe performance \citep{hewitt2019probes}, sixteen per-scenario random-label controls accompany the source-index audit. These readouts are diagnostic measurements of available associations, not inserted into the model's answer process.

Table~\ref{tab:readout-breakdown} separates the three readouts and their conjunction. The original-correct and original-error rows partition all 8,192 inputs. The last row for each model covers the errors in Section~\ref{sec:retained}: the question asks about the unchanged utterance, but the model chooses the option from the changed utterance. All three probes are correct on 488 of 617 such errors for MOSS and 206 of 218 for MiMo. The 95\% confidence intervals for these proportions are $[74.0,83.9]\%$ and $[88.9,98.6]\%$, respectively. We resample complete groups of inputs that share audio, keeping all conditions for each sampled group together.

\begin{table}[htbp]
\centering
\caption{Linear probe accuracy (\%). Q identifies the utterance asked about; S1 and S2 identify the option matching each physical utterance position. Joint requires all three predictions to be correct. These are predictions from frozen classifiers, not the decoder's answers.}
\label{tab:readout-breakdown}
\tableformat
\begin{tabularx}{\linewidth}{@{}l >{\raggedright\arraybackslash}X r r r r@{}}
\toprule
\textbf{Model} & \textbf{Original-answer population} & Q & S1 & S2 & Joint\\
\midrule
MOSS & All inputs & 93.73 & 98.78 & 98.66 & 91.53\\
& Correct & 95.08 & 99.19 & 98.91 & 93.46\\
& Incorrect & 86.41 & 96.56 & 97.27 & 81.09\\
& Unasked fact selected & 84.60 & 97.24 & 96.76 & 79.09\\
\midrule
MiMo & All inputs & 99.58 & 99.19 & 99.74 & 98.52\\
& Correct & 99.64 & 99.38 & 99.83 & 98.86\\
& Incorrect & 98.47 & 95.42 & 97.96 & 91.86\\
& Unasked fact selected & 97.25 & 98.62 & 98.62 & 94.50\\
\bottomrule
\end{tabularx}
\end{table}

Across the 16 source-index random-label controls, MOSS averages 49.976\% (range 49.011--50.598\%) and MiMo 49.946\% (48.584--50.940\%). These ranges describe seeds, not confidence intervals. These controls concern the two-class question-to-source readout, not the two four-class source-to-option tasks.

For MiMo, we additionally fit 16 random-label controls for each four-class source-to-option readout using the saved activations. Each audio group receives an independent permutation of its four labels, fixed within the group; development and test groups receive independent permutations. Each control uses the same development-group validation procedure as the true readout. Mean test accuracy is 25.15\% for the first source and 24.86\% for the second, with seed ranges of 24.32--26.48\% and 22.12--26.21\%. The corresponding true-label accuracies are 99.19\% and 99.74\% on all 8,192 inputs. These controls measure generalization under randomized labels, not the necessity of audio for decoding the true labels.

The requested physical source positions are balanced at 4,096 inputs each. Correct option counts for A--D are 1,976, 2,112, 2,056, and 2,048; the majority class occupies 25.78\%. Each physical source-to-option task has the same overall counts. Within the changed-distractor subset, correct-option counts are 848, 1,168, 1,032, and 1,048; when the changed source is asked about, they are 1,128, 944, 1,024, and 1,000. Option contents are randomly permuted per scenario and fixed within its paired conditions, rather than assigned exact equal quotas.

Reversing the order of the two utterances leaves the question, options, prompt text, and source-span layout fixed in 4,096 paired comparisons. All three readouts are correct in both orders in 85.23\% of MOSS pairs (95\% CI $[82.98,87.38]$) and 97.46\% of MiMo pairs ($[96.56,98.27]$). This tests whether decoded relations follow the utterances across positions. It is not an audio-removal control and does not establish that the decoder uses the recovered relation when choosing its answer.

A classifier that follows a physical position would give the same answer after the utterances switch places. Success in both orders instead requires the decoded option to follow the fact stated in the utterance. This makes the readouts evidence about available source associations, independently of which option the decoder chooses.

\subsection{FFN Activation Patching and Downstream Restoration}

For normalized input $h$, the gate activation $g$ scales the up-projection features $u$ element by element before the output projection:
\begin{equation}
g=\operatorname{SiLU}(W_g h),\qquad
u=W_u h,\qquad
\operatorname{FFN}(h)=W_d(g\odot u).
\label{eq:ffn}
\end{equation}

The FFN experiment transfers the activated gate $g$, the up-projection features $u$, or both from the original paired run at every option position in layers 0--17. Other activations respond normally. Channel-permutation controls rotate the donor-minus-current-receiver difference by one third of the feature dimension before adding it to the receiver, preserving its per-token Euclidean norm before numerical rounding. A normalization control uses the donor's pre-FFN RMS denominator with the receiver's current residual input and the unchanged normalization weights.

Downstream restoration uses the receiver's original option Keys, Values, or both in layers 18--35, only for post-option Query positions. Keys are taken after rotary position encoding and, where applicable, Key normalization. Restoring these attention inputs does not overwrite the evolving option hidden states. All token positions for the past utterances remain accessible.

\paragraph{Position and feature controls.}
\label{app:ffn-position-controls}
Adding 5\% of the distractor's native attention output at all option positions over all 36 layers changed the distractor-minus-correct margin by 0.059 in MOSS and 0.068 in MiMo. Adding it at question positions instead changed the margin by $-0.011$ and $-0.047$, respectively; the question-position effect varied by task in MOSS. Norm-preserving channel permutation of the gate-activation difference yielded mean directed shifts of 0.147 and $-0.222$. The activation-patching comparisons are bidirectional and do not require the contents of the two utterances to include the correct answer.

Permuting a gate change preserves its size but assigns it to different features. Its loss of directional effect shows that selecting the feature channels matters. Restoring original Keys only on later reads then tests how that earlier change reaches the answer.

\subsection{Option-Reading Transfers from Original Runs}
\label{app:option-read-method}

The independent reading experiment starts from the unedited receiver. It transfers the paired original option Keys, Values, or both solely on post-option reading connections in layers 18--35. Every option hidden state remains identical to the original receiver. The receiving Query states evolve under the intervened computation, so the experiment isolates a connection without fixing the entire subsequent trajectory.

For a single attention head and receiving Query, $C_j$ denotes the tokens of option $j$ and $a_i$ the attention on token $i$. Separate total attention weight on the options $A_C$, its shares across options $\pi_j$, and weights within an option $w_{i\mid j}$:
\begin{equation}
A_C=\sum_j\sum_{i\in C_j}a_i,\qquad
\pi_j=\frac{\sum_{i\in C_j}a_i}{A_C},\qquad
w_{i\mid j}=\frac{a_i}{\sum_{r\in C_j}a_r}.
\label{eq:reading-weights}
\end{equation}
The option contribution to this head is exactly
\begin{equation}
o_{\mathrm{opt}}=A_C\sum_j\pi_j
\underbrace{\sum_{i\in C_j}w_{i\mid j}v_i}_{\mu_j}.
\label{eq:reading-contribution}
\end{equation}
Here $v_i$ is the Value at option token $i$. This is a local identity, not an additivity assumption for final answer scores.

For the attention-allocation conditions, let $a$ be the distribution under the current Query and native receiver Keys, and let $\widetilde a$ be the distribution under that same Query and donor option Keys. Decomposing both as in Eq.~\ref{eq:reading-weights}, the between-option condition uses $A_C\widetilde\pi_j w_{i\mid j}$, the within-option condition uses $A_C\pi_j\widetilde w_{i\mid j}$, and the combined condition uses $A_C\widetilde\pi_j\widetilde w_{i\mid j}$. All retain the native non-option weights and Values. The constraints apply separately within each attention call under its current Query; they do not assert that future Queries or the full attention trajectory match the original run.

These conditions use FP32 probability calculations and add the resulting difference in weighted Values to the native attention output before its output projection. Full Key/Value transfers use native scaled-dot-product attention with the modified inputs. Self-transfers reproduce the original scores exactly. The option-prefix invariants are checked after layer 17 and at each later layer. Attention-mass and outside-weight constraint errors are approximately $1.5\times10^{-6}$; native fused BF16 attention versus explicit probability computation has a separately measured relative RMS rounding difference of approximately $0.00217$.

Fixing the total option weight leaves two ways for the weighted sum to change: shifting weight between options or shifting it between tokens within an option. Testing them separately distinguishes the choice of which option to read more heavily from the choice of which part of an option to read.

\subsection{Outcome Definitions and Uncertainty}

All main component estimates average the complete changed-distractor input population. For the original-error analysis, a qualifying pair has one original correct endpoint and one original endpoint answering with its own distractor content. Its error-production direction transfers from the incorrect endpoint to the correct endpoint. The reverse direction transfers from the correct endpoint to the incorrect endpoint. An error is reproduced only when the resulting answer label equals the incorrect endpoint's original distractor label. Source-recoverability subsets additionally require the three frozen readouts to succeed at both original endpoints.

Intervals resample the complete 512 audio-connected test groups 10,000 times, retaining all paired conditions within each sampled group. Conditional means use the eligible rows from those sampled groups. Groups with no eligible rows remain in the sampling population. Intervals are pointwise percentile intervals, not simultaneous guarantees across all conditions. The alignment across FFN, option-reading, and readout experiments uses input identities and original answer scores. The cross-experiment conjunctions and score-interaction analysis were computed after the component runs; they are not new held-out validation results.

The whole-set mean measures a directional score bias. The selected correct/error pairs measure whether that bias can produce a particular error that the unedited model actually made. Resampling whole audio groups preserves the dependence between these comparisons without treating repeated uses of an utterance as fresh evidence.

\subsection{Local Content and Weighting Effects}
\label{app:attention-derivative}

Equation~\ref{eq:content-weighting} gives the local contribution of one attention call to the final margin between a distractor and the correct option. Fix the inputs to all other calls at this layer and the Query of this call; downstream computations may respond to its output. Keys include positional transformations and any Key normalization. Let $s_i=q^\top k_i/\sqrt{d_k}$ be the score before softmax, with Key dimension $d_k$, and $a_i=\operatorname{softmax}_i(s)$. Perturb $k_i$ and $v_i$ by $\epsilon\,\delta k_i$ and $\epsilon\,\delta v_i$. The softmax derivative gives
\begin{equation}
\frac{d a_i}{d\epsilon}=a_i\left(\frac{d s_i}{d\epsilon}
-\sum_j a_j\frac{d s_j}{d\epsilon}\right),
\qquad
\frac{d s_i}{d\epsilon}=\frac{q^\top\delta k_i}{\sqrt{d_k}}.
\end{equation}
All terms are evaluated at the unperturbed attention call. Applying the product rule to $o=\sum_i a_i v_i$ yields
\begin{align}
\frac{d o}{d\epsilon}
&=\sum_i a_i\delta v_i
 +\sum_i a_i v_i\frac{d s_i}{d\epsilon}
 -\left(\sum_i a_i v_i\right)\sum_j a_j\frac{d s_j}{d\epsilon}\\
&=\sum_i a_i\delta v_i+\sum_i a_i(v_i-o)\frac{d s_i}{d\epsilon}.
\end{align}
The perturbations are zero outside $C$. Multiplying by $r^\top$, where $r=\nabla_o(\ell_d-\ell_c)$, gives Eq.~\ref{eq:content-weighting} by the chain rule. This gradient includes the output projection and the subsequent decoder, not just the output-head weights. Subtracting $o$ accounts for the change in all normalized weights; the fixed-mass experiments instead explicitly hold the non-option weights fixed.

The direction of the change at the head can also be seen without a small-perturbation assumption. Increase only $s_i$ by $t$, leaving all Values and other scores fixed. Renormalizing the weights gives exactly
\[
o(t)-o(0)=\frac{a_i(e^t-1)}{1+a_i(e^t-1)}(v_i-o(0)).
\]
For $t>0$ the head output moves toward $v_i$. This alone does not tell us which answer wins: locally, the sign for the distractor is set by $r^\top(v_i-o(0))$. These identities apply to the real-valued computation, not to rounding in the native implementation. We did not estimate $r$ or extrapolate this derivative to the finite transfers. The reported logits and choices come from actual runs, including later Query changes (Appendix~\ref{app:interactions}).

A correct probe prediction does not fix this sign. Consider a linear probe with a score difference $w^\top h$ on a state $h$, and let $g=\nabla_h m$ for the decoder's answer margin at that state. If $g$ has a component $g_\perp$ perpendicular to $w$, a small step along $g_\perp$ leaves the probe score difference unchanged, yet increases the answer margin at rate $g^\top g_\perp=\|g_\perp\|^2$. This is a compatibility argument, not a measured angle. The task does not require all scores to stay constant when an unasked fact changes. It requires the correct option to remain first. The paired errors show where that requirement fails despite correct probe predictions.

An increased attention weight favors a distractor only if the Value gaining weight moves the output in a direction favorable to that answer. The derivation specifies this local condition; the finite transfers and observed choices, not the derivative, establish its behavioral consequence.

\subsection{Recorded Runtime for the Component Experiments}

The FFN-transfer and independent option-reading runs use PyTorch 2.9.1+cu130, Transformers 4.57.1, NumPy 2.2.6, SoundFile 0.14.0, Librosa 0.11.0, and CUDA 13.0 on NVIDIA B300 SXM6 AC accelerators. TF32 is disabled. These versions are recorded consistently across the four test shards of each experiment and model; they are not inferred for the earlier behavioral or subspace experiments. The FFN-run metadata additionally records Python 3.12.3.

The checkpoint revisions are \path{d4dc5a6d8cd79b43dcd82884c75e303b1ecd016d} for MOSS-Audio-8B-Instruct and \path{2b8c0171dd106cdb709baa04c96597f53ab7ac90} for MiMo-Audio-7B-Instruct. Both use native BF16 decoder computation and causal scaled-dot-product attention. The independent option-reading runs retain the native full-shaped attention call; allocation-only interventions add FP32 probability-difference messages, as specified above. FFN projected means are saved in FP16; continuous final-margin measurements use the FP32 output head. Native discrete predictions and continuous FP32 score differences are retained separately.

A change in a continuous score difference is not itself a change in the model's predicted label. Near a tie, precision can affect which option ranks first. Keeping native choices separate from higher-precision margins preserves the distinction between the decision and its estimated size.

The methods separate three questions: which association is available, what a later token receives from an option, and whether the resulting score change alters the choice. Holding the receiver and its options fixed connects these measurements without treating probe accuracy or attention weight as the answer itself.

\section{Additional Results for the Option-Level Interventions}
\label{app:component-results}

\subsection{Complete Component Comparisons}

Table~\ref{tab:component-full} expands Figures~\ref{fig:gate-reading} and~\ref{fig:reading-components}. It includes the up-projection, joint gate/up, channel-permutation, and normalization controls. All estimates use the complete changed-distractor population, with transfers defined by the paired materials before answer correctness is examined. The gate and up branches differ across the models: transferring the up branch alone has a negative mean directed effect in MOSS but a positive one in MiMo. Thus the gate experiment identifies a content-dependent transformation and its downstream path, rather than establishing an exclusive FFN mechanism.

\begin{table}[htbp]
\centering
\caption{Full directed-score comparisons for the changed-distractor inputs ($n=4,096$ per model). Entries are means and 95\% intervals from 10,000 complete audio-group resamples. FFN transfers act on option positions at layers 0--17; reading changes act on post-option Queries at layers 18--35.}
\label{tab:component-full}
\tableformat
\begin{tabularx}{\linewidth}{@{}>{\raggedright\arraybackslash}X r c r c@{}}
\toprule
 & \multicolumn{2}{c}{\textbf{MOSS}} & \multicolumn{2}{c}{\textbf{MiMo}}\\
\cmidrule(lr){2-3}\cmidrule(l){4-5}
\textbf{Intervention} & $\Delta s$ & 95\% CI & $\Delta s$ & 95\% CI\\
\midrule
Early gate & $3.737$ & $[3.570,3.907]$ & $1.598$ & $[1.497,1.700]$\\
Early up & $-0.248$ & $[-0.372,-0.121]$ & $0.623$ & $[0.546,0.701]$\\
Early gate and up & $3.360$ & $[3.195,3.523]$ & $1.158$ & $[1.075,1.242]$\\
Permuted gate difference & $0.147$ & $[0.123,0.172]$ & $-0.222$ & $[-0.245,-0.200]$\\
Permuted up difference & $-0.108$ & $[-0.134,-0.082]$ & $0.128$ & $[0.098,0.157]$\\
Normalization control & $-0.016$ & $[-0.019,-0.012]$ & $0.032$ & $[0.027,0.038]$\\
\midrule
Gate + restore late Keys & $0.452$ & $[0.421,0.483]$ & $0.322$ & $[0.299,0.346]$\\
Gate + restore late Values & $3.326$ & $[3.176,3.478]$ & $1.381$ & $[1.285,1.477]$\\
Gate + restore both & $0.017$ & $[0.014,0.019]$ & $0.028$ & $[0.024,0.031]$\\
\midrule
Transfer option Keys & $4.933$ & $[4.748,5.119]$ & $2.093$ & $[1.959,2.230]$\\
Transfer option Values & $0.632$ & $[0.589,0.674]$ & $0.492$ & $[0.455,0.530]$\\
Transfer option Keys and Values & $5.564$ & $[5.343,5.784]$ & $2.573$ & $[2.417,2.733]$\\
\midrule
Between-option allocation & $3.118$ & $[2.979,3.259]$ & $1.585$ & $[1.476,1.698]$\\
Within-option allocation & $1.830$ & $[1.732,1.931]$ & $0.574$ & $[0.529,0.621]$\\
Both allocations (fixed mass) & $4.860$ & $[4.675,5.045]$ & $2.025$ & $[1.896,2.157]$\\
\bottomrule
\end{tabularx}
\end{table}

\FloatBarrier

The gate transfer favors the donor's distractor in both models, whereas the up-projection alone does not even have the same sign across models. The common finding is that a particular reweighting of features can bias later option reads, not that changing any FFN activation has the same effect.

\subsection{Relevant Use and Irrelevant Interference}
\label{app:relevant-use}

We evaluate changes to the same utterance under both questions. Each role has 4,096 inputs and 2,048 undirected content pairs. When the question asks about the changed fact, correct answers should update with its content. When the question asks about the unchanged fact, correct answers should remain the same. Table~\ref{tab:relevant-behavior} separates single-input accuracy, paired correctness, and following the changed content. Stability alone would count two identical wrong answers as success, so the primary stability measure requires both answers to be correct.

\begin{table}[htbp]
\centering
\caption{Original model behavior under both question roles. Percentages and 95\% intervals use 10,000 resamples of all 512 complete audio groups. Single-input metrics use 4,096 inputs per role; update and stability metrics use 2,048 content pairs per role.}
\label{tab:relevant-behavior}
\tableformat
\begin{tabularx}{\linewidth}{@{}>{\raggedright\arraybackslash}X c c@{}}
\toprule
\textbf{Outcome} & \textbf{MOSS} & \textbf{MiMo}\\
\midrule
Asked: single-input accuracy & $83.94\;[82.30,85.57]$ & $96.17\;[95.39,96.90]$\\
Asked: both correct and updated & $76.51\;[74.32,78.76]$ & $93.31\;[91.99,94.58]$\\
\midrule
Unasked: single-input accuracy & $84.81\;[83.03,86.65]$ & $94.24\;[92.97,95.41]$\\
Unasked: both correct and stable & $78.12\;[75.73,80.57]$ & $91.06\;[89.40,92.63]$\\
Unasked: changed content selected & $15.06\;[13.26,16.85]$ & $5.32\;[4.25,6.47]$\\
\bottomrule
\end{tabularx}
\end{table}

For the intervention comparison, Eq.~\ref{eq:directed-score} always measures the donor-content-minus-receiver-content shift. When the question concerns the changed utterance, these contents are the respective correct answers of the two original inputs. A shift toward the donor content is therefore evidence of content transmission, not an accuracy loss measured against the receiver's unchanged label. When the question concerns the other utterance, both contents are incorrect options. Table~\ref{tab:relevance-difference} compares the same content pair, utterance order, and intervention across these question roles. The larger effects in the asked condition show relevance-dependent use of this path; the nonzero unasked effects quantify its influence on irrelevant content. Changing the question also changes its semantics, so this is a paired task comparison rather than an isolated manipulation of one internal relevance variable.

\begin{table}[htbp]
\centering
\caption{Asked-minus-unasked difference in directed score shifts from Table~\ref{tab:asked-unasked-path}. Entries are paired means with pointwise 95\% complete-audio-group intervals, not accuracy differences.}
\label{tab:relevance-difference}
\tableformat
\begin{tabularx}{\linewidth}{@{}>{\raggedright\arraybackslash}X c c@{}}
\toprule
\textbf{Transferred activation} & \textbf{MOSS} & \textbf{MiMo}\\
\midrule
Early FFN gate & $3.635\;[3.412,3.859]$ & $5.025\;[4.836,5.213]$\\
Later option Keys & $4.587\;[4.362,4.817]$ & $7.540\;[7.338,7.750]$\\
Later option Values & $1.445\;[1.377,1.515]$ & $1.698\;[1.646,1.751]$\\
Later option Keys and Values & $6.057\;[5.772,6.350]$ & $9.291\;[9.067,9.524]$\\
\bottomrule
\end{tabularx}
\end{table}

A question changes how much a fact from a past utterance counts. All four interventions have larger mean effects when the fact is requested. Their positive mean effects when it is not requested show that the same computation that supports a correct update can also favor an irrelevant answer.

\subsection{Joint Score Effects and Discrete Errors}
\label{app:interactions}

For a fixed Query in one attention head, let $a_i$ and $a'_i$ be the weights on option token $i$ with the original and transferred Keys, respectively, and $\Delta v_i$ the change in its Value. Subtracting the two single-transfer changes from the joint change in the head output gives $\sum_{i\in C}(a'_i-a_i)\Delta v_i$, where $C$ contains the option tokens. The remainder vanishes if weights stay fixed wherever Values change. This identity and Eq.~\ref{eq:reading-contribution} concern one read; later Queries may change. We quantify the interaction in final option scores by
\begin{equation}
I_{K,V}=\Delta s_{K,V}-\Delta s_K-\Delta s_V,
\end{equation}
where every $\Delta s$ subtracts the same original-run margin as in Eq.~\ref{eq:directed-score}. The joint contrast is therefore equivalent to the joint margin minus the two single-transfer margins plus the original margin.

\begin{figure}[htbp]
\centering
\includegraphics[width=\linewidth]{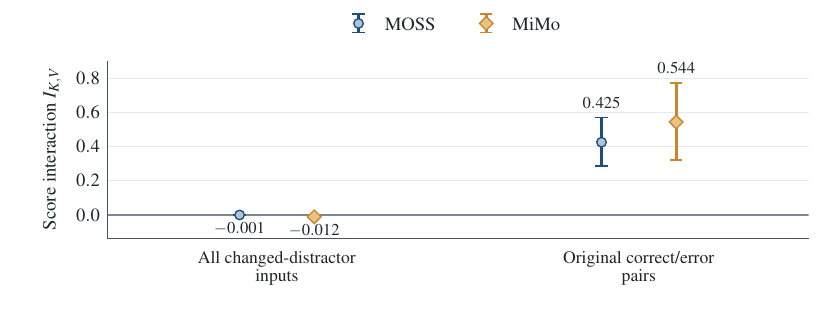}
\caption{Key/Value score interaction depends on the analysis population. The left group uses all 4,096 changed-distractor inputs per model; the right group uses 273 MOSS and 123 MiMo original correct/error pairs. Points are means; whiskers are pointwise 95\% intervals from resampling all 512 audio groups, including groups with no eligible error pairs. These are score interactions, not accuracy gains.}
\label{fig:score-interaction}
\end{figure}

Across all 4,096 changed-distractor inputs, MOSS has mean interaction $-0.00108$, with a 95\% interval of $[-0.00510,0.00298]$. MiMo has mean $-0.01187$, with interval $[-0.01693,-0.00691]$. In the original correct/error pairs from Section~\ref{sec:natural-errors}, the means are $0.4254$ ($[0.2861,0.5685]$) and $0.5438$ ($[0.3209,0.7722]$). Figure~\ref{fig:score-interaction} contrasts these populations. The cohort average is close to additive in scale, but it does not characterize the error subset or establish per-input additivity.

The discrete outcome supplies a separate test. Joint Key/Value transfer produces the specific original error when neither single transfer does in 133/273 MOSS pairs (48.7\%, 95\% CI $[42.0,55.4]$) and 48/123 MiMo pairs (39.0\%, $[30.6,47.9]$). Likewise, jointly changing between-option and within-option allocation produces the error without either single allocation change doing so in 87/273 and 26/123 pairs. A continuous score change and crossing an answer decision boundary are distinct events. These option-reading results are not the earlier joint intervention on two past utterances; they concern different variables and should not be treated as the same additivity test.

On the same original correct/distractor-error pairs, patching the early FFN gate activations alone produced the specific paired distractor in 51/273 MOSS pairs and 11/123 MiMo pairs. Joint later option Key/Value patching did so in 87.9\% (95\% interval $[83.5,91.8]$) and 70.7\% ($[62.2,79.0]$), respectively. The joint later patch produces this specific wrong answer in more pairs.

\begin{table}[htbp]
\centering\small\tableformat
\caption{Score changes behind Table~\ref{tab:specific-errors}. Top: mean final logit margin $m=\ell_d-\ell_c$ on the same pairs; positive means that the original distractor $d$ outscores the correct option $c$. Bottom: checks on transfers into the originally correct run.}
\label{tab:paired-error-details}
\begin{tabular*}{\linewidth}{@{}l@{\extracolsep{\fill}}rrrr@{}}
\toprule
& \multicolumn{2}{c}{Originally wrong input} & \multicolumn{2}{c}{Originally correct input}\\
\cmidrule(lr){2-3}\cmidrule(l){4-5}
Transferred component & MOSS & MiMo & MOSS & MiMo\\
\midrule
None & $+1.03$ & $+1.23$ & $-5.00$ & $-4.47$\\
Early FFN gate & $-3.04$ & $-2.39$ & $-1.43$ & $-2.52$\\
Option Keys & $-4.36$ & $-3.18$ & $-0.12$ & $-0.73$\\
Option Values & $+0.01$ & $-0.19$ & $-4.47$ & $-3.59$\\
Option Keys and Values & $-4.91$ & $-4.11$ & $+0.86$ & $+0.40$\\
\bottomrule
\end{tabular*}
\smallskip
\begin{tabularx}{\linewidth}{@{}>{\raggedright\arraybackslash}Xrr@{}}
\toprule
Joint Key/Value transfer & MOSS & MiMo\\
\midrule
Extra change in $d$ versus gate alone, relative to the unmentioned option (logits) & $+1.86$ & $+2.15$\\
Extra change in $c$ on the same scale (logits) & $-0.43$ & $-0.77$\\
Joint transfer selects $d$ and neither single transfer does (pairs) & 133/273 & 48/123\\
Of those pairs, sum of single-transfer score changes predicts $d$ & 61/133 & 35/48\\
Joint transfer selects $d$ among pairs with all three probes correct in both runs & 190/215 & 83/118\\
\bottomrule
\end{tabularx}
\end{table}

\paragraph{Why the gate transfer is asymmetric.}
For Table~\ref{tab:specific-errors} we reanalyzed the four saved option logits, keeping the same wrong option $d$ and correct option $c$ in both directions. Margins use the FP32 output projection; choice counts use the original model-precision argmax over all four options. We resampled all 512 audio groups 10,000 times, retaining both directions of every pair and groups with no qualifying pairs. The correct answer's original lead over $d$ exceeded $d$'s lead in the paired wrong run in 271/273 MOSS pairs and 120/123 MiMo pairs. Transferring the wrong run's early gates raised $m=\ell_d-\ell_c$ by 3.56 and 1.94 logits on average. Restoring the correct run's later option Keys reduced these shifts to 0.43 and 0.42; restoring Values instead left shifts of 3.06 and 1.70.

\paragraph{The distractor's gain relative to an unmentioned option.}
Relative to the gate transfer, joint Key/Value transfer raised $m$ by 2.29 logits in MOSS (95\% interval $[2.07,2.52]$) and 2.92 in MiMo ($[2.59,3.28]$). Using the unmentioned fourth option as a common reference, the distractor rose by 1.86 and 2.15 logits, while the correct option fell by 0.43 and 0.77. These within-pair comparisons test two different interventions; their magnitudes were not matched to rank the importance of the FFN and attention.

To check whether the single replacements predict the joint choice, we added their shifts for each of the four options: $\widehat{\ell}_{K,V}=\ell_K+\ell_V-\ell_{\rm original}$. With the stored model-precision logits, this predicts $d$ on 61/133 MOSS and 35/48 MiMo pairs where only joint replacement selects $d$. Keeping the original pairs fixed and using FP32 to score all post-intervention choices gives 64/130 and 37/49. Two individually insufficient shifts can thus account for some joint choices; their sum does not account for all of them.

A small average interaction across all inputs can coexist with a larger interaction on pairs that change winners. Moreover, even exactly additive score shifts can cross the decision boundary together. A wrong choice produced only by the joint transfer is therefore not, by itself, evidence of a nonlinear interaction.

\subsection{Training an Adapter to Combine Audio Clips in Early Layers}
\label{app:early-fusion-training}

This development experiment trained an adapter, not the original FFN gates. With the backbone frozen, the adapter maintained a separate state for each audio clip in layers 12--17 and combined the states at layer 17. An ordinary residual adapter at layer 17 served as a control with a similar parameter count. Both trained on the same 3,030 pairs for one epoch. Neither received the index of the relevant clip at inference.

We evaluated all 517 MMAR-derived questions with two, three and four audio clips per question. We rescored the frozen model under the same inference settings as both adapters and counted correct greedy first tokens. Across these three conditions, mean accuracy for the frozen model, the per-clip adapter and the ordinary adapter was 54.09\%, 55.32\% and 55.25\% in MOSS, and 53.51\%, 54.80\% and 58.67\% in MiMo. Keeping separate states gave no consistent advantage over ordinary adaptation in this single-seed comparison. These are training results on a different task from the gate replacements, not estimates of the fraction of errors caused by the FFN.

Keeping the clips' states separate until fusion improved accuracy over the frozen model but did not consistently outperform an ordinary adapter. On this comparison, reserving a state for each audio clip is not enough to explain the improvement: those states still have to be weighted appropriately for the question.

The shared result across models is not a particular adapter or a universal additive law. The question determines how strongly a fact from a past utterance influences the answer, and in the original correct/error pairs the later option read can give the irrelevant fact enough support to win. These are properties of the measured computation and its decision boundary.

\section{Behavioral Materials and Answer-Format Details}
\label{app:behavior-methods}

\subsection{Interference from Distractor Utterances in LibriSpeech}
\label{app:librispeech-interference}

We assemble 1,000 dialogue histories from LibriSpeech utterances: 450 use test-clean and 550 use test-other. Each history contains utterances from different speakers and chapters. Each of the three questions concerns a different utterance, giving 3,000 questions. The neutral condition substitutes utterances without the option answer words for the non-target utterances, matching word count and duration. The transcript comparison replaces utterances with their reference text and keeps the question and options unchanged.

The audio accuracy changes in Section~\ref{sec:behavior} are from 84.23\% to 78.73\% for MOSS and from 48.23\% to 39.67\% for MiMo. With reference text, the corresponding changes are 98.10\% to 96.57\% and 70.43\% to 52.50\%. These are modality-specific effects, not a claim that audio invariably causes stronger interference. The relevant-only comparison uses the intersection of 1,365 questions both models answer correctly from the relevant utterance alone. We average the paired accuracy change within each of the 830 dialogue histories represented in this subset, then weight them equally.

\begin{table}[htbp]
\centering\small\tableformat
\caption{LibriSpeech utterances: accuracy levels and paired differences. Only the neutral utterances are replaced; the relevant utterance stays fixed. Score coefficients are changes in the designated wrong-minus-correct logit difference per additional supporting turn.}
\label{tab:behavior-estimates}
\begin{tabularx}{\linewidth}{@{}>{\raggedright\arraybackslash}Xcc@{}}
\toprule
Measure & MOSS & MiMo \\
\midrule
Neutral dialogue history: accuracy (\%) & 84.23 & 48.23 \\
Competing dialogue history: accuracy (\%) & 78.73 & 39.67 \\
Paired accuracy change (pp), 95\% CI & $-5.50\;[-6.73,-4.30]$ & $-8.57\;[-9.80,-7.37]$ \\
Change on relevant-only-correct items (pp) & $-2.79$ & $-10.06$ \\
\midrule
Nine turns: support for designated wrong option & $0.260$ & $0.089$ \\
Nine turns: support for another wrong option & $0.008$ & $-0.001$ \\
95\% CI for the coefficient difference & $[0.195,0.305]$ & $[0.068,0.111]$ \\
\bottomrule
\end{tabularx}
\end{table}

The nine-turn experiment uses 300 dialogue histories. The relevant utterance appears at position 1, 5, or 9 in 100 histories each. Zero, one, two, four, or eight other utterances support a specified incorrect option. Increasing this count replaces neutral utterances rather than appending turns. Accuracy changes from 84.67\% to 65.67\% in MOSS and from 48.33\% to 39.67\% in MiMo between zero and eight supporting distractors. The directional-score comparison distinguishes support for the particular competing option from support for a different incorrect option.

With the number of utterances fixed, a mention favors the wrong option it names much more than a different wrong option. The text comparison shows why this is not simply an acoustic penalty: the direction is shared across modalities, whereas the larger accuracy drop occurs in audio for one model and in text for the other.

\subsection{Prompt Sequence and Case Selection}

The controlled comparisons use the system instruction, ``You are a helpful assistant. Answer the question using the provided information.'' Each audio turn is introduced with ``Please listen to this recording.'' and followed by the fixed assistant acknowledgement ``Understood.'' The final text turn presents the question and, for multiple-choice QA, the four labeled options followed by ``Respond with only the letter of the selected option.'' The model's native prompt wrapper is retained. The MiMo prompt includes its empty thinking block before scoring the answer.

Open-ended QA removes the displayed options and uses ``Answer briefly with the value only.'' Greedy generation is capped at 32 new tokens. Multiple-choice scoring instead compares the next-token logits of A--D; it does not parse a generated explanation. These measurements must not be combined as though they were the same output protocol.

The cases in Table~\ref{tab:recorded-cases} use options A: purple, B: black, C: orange, D: red for MOSS, and A: box, B: locker, C: suitcase, D: drawer for MiMo. They were selected in input-identifier order among same-family cases with correct relevant-only answers, a distractor error with the full dialogue history, and correct answers when asked about the event in the distractor utterance. They are qualitative examples, not an estimate of error prevalence. The case in Table~\ref{tab:paired-example} illustrates the content-pair design; Figure~\ref{fig:overview}a shows the observed MOSS example from the case table, while its right panel summarizes the mechanism experiments.

The identical acknowledgements contribute no answer words. A change of question can therefore make the previously wrong answer correct without any change in the earlier dialogue. The selected cases illustrate that distinction; the full paired sample, not the cases, measures its frequency.

\subsection{Denominators for Blocking and State Transfer}

The strict-block test has 512 scenarios and four content/question conditions each; the distractor utterance precedes the relevant utterance. With the full dialogue history, the distractor error counts are 175 for MOSS and 94 for MiMo. Following just these originally erroneous conditions yields the final two columns of Table~\ref{tab:access-states}. Across all 2,048 conditions, the strict late block produces 131 and 57 distractor answers, including 7 and 8 newly introduced ones. Replacing with early-blocked states gives totals of 6 and 2; restoring full-access states gives 132 and 58. The original-error subset and the whole-population counts answer different questions and are not interchangeable. The option-level interventions in Section~\ref{sec:mechanism} separately include both utterance orders.

\begin{figure}[htbp]
\centering
\includegraphics[width=\linewidth]{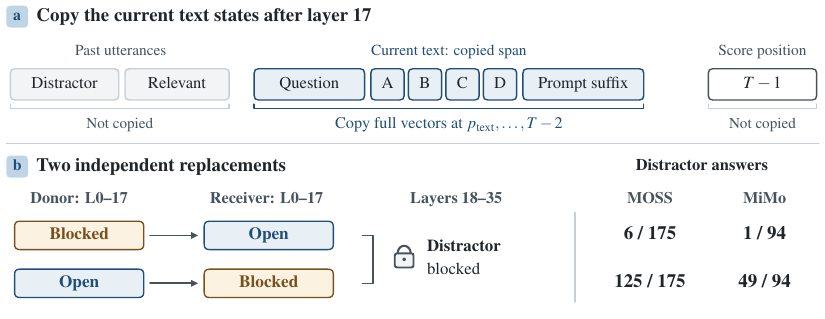}
\caption{State transfer with later access to the distractor blocked. (a) After layer 17, copy the full hidden vectors from the current user-role prefix through the penultimate prompt token; neither the audio nor the answer-scoring position is copied. (b) Open and blocked refer to early access from the text turn to the distractor. Each arrow is a separate donor-to-receiver replacement under the same strict late block. Counts track the original multiple-choice errors. Without replacement the late block leaves 124/175 in MOSS and 49/94 in MiMo.}
\label{fig:state-bridge}
\end{figure}

Counting the original errors again measures persistence; counting all errors after the block measures the error rate under that condition. The latter includes newly induced errors. Keeping the two denominators distinct shows that a change in the total is not the same as repairing that many individual answers.

\subsection{An Oracle Source Control and Uncued Training Details}
\label{app:source-use-validation}

\paragraph{Supplying the correct source as a diagnostic control.}
All conditions use the same 512 test scenarios, including both utterance orders, question roles and versions of the changed fact (8,192 inputs per model). Tables~\ref{tab:uncued-source-use}a and~\ref{tab:source-confirmation} report baselines from their own evaluations, not the original predictions in Table~\ref{tab:relevant-behavior}. We use the baseline from the same run protocol for each comparison. Unlike studies that elicit a clarification from a user or evaluate user preferences for repair strategies \citep{zhang2025clarifying,ashktorab2019resilient}, this control supplies an oracle source index. We prepend \emph{Use only the [first/second] recording to answer the question below.} The model receives neither the answer nor states from another run. No earlier wrong reply or human correction turn is included. Both restricted conditions block direct attention to the token span of the unselected utterance at all 36 decoder layers; they differ only in which current-turn positions are subject to the block. For single-utterance replay, we send a fresh request containing only the relevant utterance, the original question and the options. Replay also changes input length and positions.

\begin{table}[htbp]
\centering\small\tableformat
\caption{Using the confirmed source in controlled multiple-choice QA. A distractor answer selects the option named in the distractor utterance. Counts use all 8,192 inputs per model. The three instruction conditions have identical prompts.}
\label{tab:source-confirmation}
\begin{tabular}{@{}lrrrr@{}}
\toprule
& \multicolumn{2}{c}{Accuracy (\%)} & \multicolumn{2}{c}{Distractor answers} \\
\cmidrule(lr){2-3}\cmidrule(l){4-5}
Condition & MOSS & MiMo & MOSS & MiMo \\
\midrule
Full dialogue history, no source cue         & 84.12 & 95.31 & 1,288 & 357 \\
Source instruction alone            & 93.07 & 98.73 & 554 & 57 \\
Instruction + answer-position block & 93.26 & 98.74 & 538 & 55 \\
Instruction + current-turn block    & 99.37 & 99.15 & 23 & 17 \\
Fresh request, selected utterance only  & 99.41 & 99.29 & 20 & 12 \\
\bottomrule
\end{tabular}
\end{table}

Compared with the source instruction alone, blocking the unselected utterance throughout the current turn increased accuracy by 6.30 percentage points in MOSS (95\% CI: 5.59 to 7.02) and 0.42 points in MiMo (0.09 to 0.73). Intervals use 2,000 paired bootstrap draws over the 512 complete audio-connected groups, retaining all input variants and conditions within a sampled group.

\paragraph{Answer supervision without a source cue.}
For each training scenario we change a fact in one utterance and ask about each of the two sources in turn, yielding four inputs with the same options. In one pair the correct answer changes with the fact; in the other it stays the same. We fit LoRA adapters of rank 8 \citep{hu2022lora} to the attention and FFN projections in all 36 decoder layers on 128 scenarios, with no audio shared with the 512 test scenarios. We compare ordinary four-option cross-entropy with the same loss plus a Jensen--Shannon penalty (weight 1) between the two answer distributions whose correct answer is unchanged. Both conditions use identical inputs, 384 updates and 3 matched seeds; we evaluate the final checkpoints. With ordinary supervision, the answers updated correctly on 94.92\% of the relevant-change pairs in MOSS and 99.54\% in MiMo; they stayed correct on 95.02\% and 99.67\% of the irrelevant-change pairs. Each measure requires both endpoints to be correct. Adding the penalty changed overall accuracy by $-0.18$ and $+0.02$ percentage points, respectively, averaged over these seeds. This comparison isolates the extra penalty, not the benefit of pairing over training on an equally large unpaired dataset.

A spoken correction within a single conversation is a different task: the assistant must also track which version of an event the question asks about \citep{wu2025longmemeval}. The controlled training and test here use alternative dialogue histories, not successive corrections.

With the correct source supplied, blocking the distractor throughout the question turn is more effective than blocking only the answer position. In the uncued test, ordinary answer supervision on both question roles already improves correct updates and correct stability. Neither result requires treating a user-supplied utterance index as a practical interface.

\subsection{Adapting Reads from Audio Clips during the Question Turn}
\label{app:value-read-adaptation}

\paragraph{What is trained.}
For each model we froze the backbone and added a rank-32 module in layers 9--17 (195,840 trainable parameters). On 22 separate development scenarios, we selected eight attention heads per layer for each model independently. At each prompt position from the start of the question turn through the answer prompt, the module uses the current attention Query and each earlier turn's attention-weighted sum of Values to compute a residual at the head output. Each audio turn includes the clip, the chat delimiters and the acknowledgement. The module does not replace the local attention weights, erase an audio clip or require its relevance label. Each model was trained once, using the same seed and the same ordered 4,848 pairs: 1,600 controlled temporal QA pairs, 350 pairs from MMSU \citep{wang2026mmsu}, and 348 distinct MMAR pairs repeated to give 2,898 presentations. We used 1,212 AdamW updates with full-vocabulary answer cross-entropy and a paired loss (weight 0.2): it favors agreement when the gold answer is unchanged and a change in the answer margin when the gold answer changes.

\paragraph{Evaluation and overall accuracy.}
We used the original questions and options from 517 MMAR items \citep{ma2025mmar} whose parent videos were disjoint from training. Each question was evaluated with the target audio clip alone and with one, two or three additional clips; the prompt never identified the target. We scored the greedy first token over the full vocabulary against the correct option label. The 1,551 responses to questions with multiple audio clips give the overall accuracy in Section~\ref{sec:implications}. Its paired confidence interval uses 10,000 bootstrap draws over 115 complete parent-video groups, retaining all questions and clip counts in each draw. These results describe our adaptation of MMAR to questions with multiple clips, not the official benchmark.

MiMo improved by 5.80, 6.96 and 5.61 percentage points with two, three and four audio clips, respectively. The gain was not solely a change in output format: scoring the highest-logit option instead of the greedy first token still gave a 5.48-point gain (53.97\% to 59.45\%). With only the target audio clip, accuracy also rose, from 56.67\% to 60.15\% in MOSS and from 56.87\% to 62.48\% in MiMo. As a secondary comparison, we subtracted the accuracy gain with one audio clip from the gain with multiple clips on the same questions. The differences were $+1.16$ percentage points in MOSS (paired 95\% CI $[-1.80,4.07]$) and $+0.52$ in MiMo ($[-2.79,3.88]$), using the same parent-video bootstrap. Both intervals include zero, so this comparison does not establish a larger gain with distractor clips present.

The frozen-backbone adjustment improves both models' accuracy on questions with multiple audio clips without requiring a source label. Because it also improves accuracy with a single clip, the overall gain is useful evidence for modifying the read, but the experiment cannot attribute that gain specifically to suppressing interference.

The original answers, the blocking controls and the training experiments answer different practical questions. The first identify the particular fact used in error; the second show when limiting access still changes that answer. The training evaluations measure whether answers improve without asking the user to identify the source.

\section{Construction-Order and Question-Wording Controls}
\label{app:subspace-controls}

Switching a question from an original plan to its update changes both the requested source and the answer. We test whether the corresponding changes in hidden states can nevertheless be attributed separately to the question and to what the utterances say. A question contrast changes the requested arrangement while keeping the utterances fixed; a content contrast changes the fact in the original utterance while keeping the question about the updated arrangement fixed. We call the resulting directions \emph{question-contrast} and \emph{content-contrast} directions. The fact stated in an utterance here is an answer value such as a time or color, whereas the attention Value in Section~\ref{sec:candidate-reading} is a vector used in the attention computation.

\subsection{Protocol and Construction Order}
\label{app:subspace-order}

The materials contain an original arrangement and an updated arrangement for the same person, such as a departure time. They use 128 development and 512 test scenarios, spanning times, colors, quantities, and locations. The 512 test scenarios form 512 complete audio-connected groups. These are previously examined test materials, not a new unseen confirmation set. Donors are assigned from development scenarios in the same family before test scoring. All bases are fitted on development data with the original question templates.

Question-first construction uses the question contrast as the reference and tests whether a content edit outside its fitted directions further changes the answer. At each layer, we first fit a rank-four principal-component basis to question differences, remove that projection from the content differences, and fit the second rank-four basis to the residual. Content-first construction reverses these operations and changes the reference question edit. The fitting order defines this conditional comparison, not a processing order inside the model. Differences are computed in native BF16, then centered and projected in FP32; randomized SVD uses seven iterations and a fixed seed. A projected development-donor increment is added at one layer to the final 32 prompt positions, including the answer-scoring position. The test covers each of the 36 layers, both construction orders, and question-only, content-only, and joint increments. Unlike Section~\ref{sec:mechanism}, these are single-layer suffix interventions, not transfers confined to option-reading connections.

\paragraph{Why reversing construction changes the experiment.}
Let $D_Q$ and $D_C$ contain the development question and content differences, with one row per token state, and let $\operatorname{PCA}_4$ return four orthonormal directions. Question-first construction uses
\begin{equation}
U_Q=\operatorname{PCA}_4(D_Q),\qquad
U_C=\operatorname{PCA}_4\!\left(D_C(I-U_QU_Q^\top)\right).
\label{eq:subspace-construction}
\end{equation}
Content-first construction starts with $\widetilde U_C=\operatorname{PCA}_4(D_C)$ and fits $\widetilde U_Q$ to the remaining question differences. PCA retains directions of high development variance, not directions chosen for their effect on the answer. When the question changes from original to updated, both the requested event and its answer change. The resulting state difference need not be specific to the source. Fitting content first excludes the leading content directions from the question fit; it does not remove only changes in wording or only changes in answer content.

Two things can therefore change. Write $P_Q^O,P_C^O$ for the question and content projectors in fit $O$, and $P^O=P_Q^O+P_C^O$ for their joint projector. For one token let $d_Q$ be the donor's measured change when switching the question, and $d_C$ its change when switching the original fact at fixed question. Let $A$ denote question-first and $B$ content-first. Before rounding, the requested joint edit is $e_O=P_Q^O d_Q+P_C^O d_C$. Subtraction gives the exact identity
\begin{equation}
e_B-e_A=(P^B-P^A)d_C+(P_Q^B-P_Q^A)(d_Q-d_C).
\label{eq:subspace-order-difference}
\end{equation}
The first term allows the joint span to change: the two lower-rank fits need not retain the same directions. The second applies even if the joint span stays fixed. Along a direction reassigned from question to content, the coefficient now comes from the measured change in the fact stated in the utterance, not the measured change in the question. If those coefficients agree and the joint span is unchanged, reassigning the direction has no effect. Simply reversing a basis vector's sign never changes its projector.

At layer 23 the largest cosine between the independently fitted question and content spaces is 0.499 in MOSS and 0.816 in MiMo. Across orders the two joint spans have overlap 0.984 and 0.852, respectively (mean squared cosine of their principal angles). In MOSS, the joint span is nearly unchanged, while the joint margin shift is 2.719 under question-first fitting and 2.663 under content-first fitting. Span overlap describes the retained directions, not the donor coefficients assigned within them or their effect on an answer. We did not vary the two terms of Eq.~\ref{eq:subspace-order-difference} independently.

\paragraph{How the change reaches the answer scores.}
For question-only edits, $e_{Q,O}=P_Q^O d_Q$, giving $e_{Q,B}-e_{Q,A}=(P_Q^B-P_Q^A)d_Q$. Unlike the joint comparison in Eq.~\ref{eq:subspace-order-difference}, this changes only the question projector. Let $z_A,z_B$ be the final output-head inputs for the \emph{same test question} under these edits, and $w_u,w_o$ the output weights for the updated and original answer labels. The linear output head gives the margin $m=\ell_u-\ell_o$ and difference
\begin{equation}
m_B-m_A=(w_u-w_o)^\top(z_B-z_A).
\label{eq:order-answer-readout}
\end{equation}
Thus even a large change in the final state orthogonal to $w_u-w_o$ would leave this comparison unchanged. With the question edit fitted second, the remaining decoder produces a smaller shift along this answer comparison. The FP32 output-head shifts at layer 23 go from 2.149 to 1.995 in MOSS and from 0.815 to 0.231 in MiMo. These are finite changes in the updated-minus-original margin, not separate measurements of the two logits or a localization to a particular head. In particular, a subspace intervention can move an answer without uniquely identifying the feature used in the original run \citep{makelov2024subspace}.

\begin{figure}[htbp]
\centering
\includegraphics[width=\linewidth]{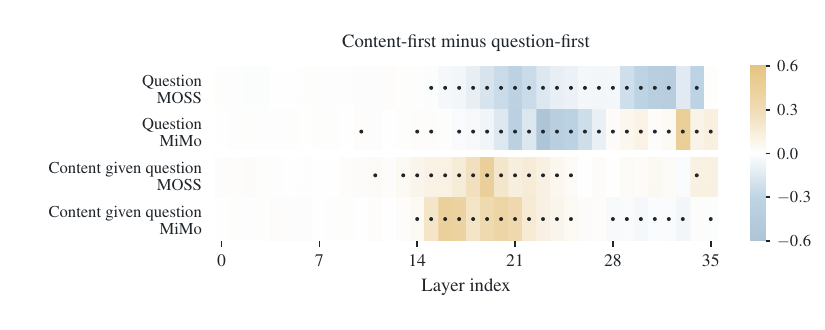}
\caption{Construction order affects subspace interventions. Cells show content-first minus question-first effects on the answer margin at all 36 tested layers; blue is negative and gold positive. The conditional content effect is joint minus question-only intervention. Dots mark cells whose pointwise 95\% interval excludes zero, using 10,000 resamples of all 512 complete audio groups, without a multiple-comparison correction. Native next-token scores are used. These are unscaled projection comparisons, not norm-matched interventions.}
\label{fig:subspace-order}
\end{figure}

The receiver asks about the original event, so a positive updated-minus-original margin shift is not an accuracy improvement. Figure~\ref{fig:subspace-order} shows order effects on the question-only shift and the additional content effect. All increment multipliers are one; projected magnitudes need not match across orders.

For example, at layer 23 the question-contrast effects are 2.154 versus 1.993 in MOSS and 0.811 versus 0.234 in MiMo. The paired order changes are $-0.161$ (95\% CI $[-0.206,-0.117]$) and $-0.577$ ($[-0.641,-0.514]$). The actual mean RMS increments also change, from 0.319 to 0.301 in MOSS and from 0.080 to 0.070 in MiMo. Because the downstream computation is nonlinear, dividing effects by these mean magnitudes would not substitute for a matched-amplitude intervention.

\paragraph{The content effect shared by both constructions.}
Within each construction we subtract the question-only shift from the joint shift. This measures what adding the content increment contributes at the same layer, not a sequence of processing stages. Table~\ref{tab:conditional-content-order} reports both orders at the five prespecified layers; every pointwise interval is above zero. At layer 23, fitting content first increases this conditional effect by 0.105 logits in MOSS (95\% CI $[0.052,0.159]$) and 0.106 in MiMo ($[0.062,0.151]$). That gain offsets most of the 0.161 decrease in the MOSS question-only effect but little of the 0.577 decrease in MiMo: the joint shifts change from 2.719 to 2.663 and from 1.215 to 0.744, respectively. This is an arithmetic comparison of measured margins, not a claim that the separate edits add linearly. What persists under both fits is the contribution of what the utterance says over and above the question edit, not a unique division of the answer preference into source and content.

\begin{table}[htbp]
\centering
\caption{Additional content effects at the five prespecified layers. Each entry is the joint minus question-only logit-margin shift, with a pointwise 95\% interval over 512 complete audio groups. Both constructions use their original projection magnitudes; these are not amplitude-matched comparisons.}
\label{tab:conditional-content-order}
\tableformat
\begin{tabularx}{\linewidth}{@{}r *{4}{>{\centering\arraybackslash}X}@{}}
\toprule
 & \multicolumn{2}{c}{\textbf{MOSS}} & \multicolumn{2}{c}{\textbf{MiMo}}\\
\cmidrule(lr){2-3}\cmidrule(lr){4-5}
\textbf{Layer} & Question-first & Content-first & Question-first & Content-first\\
\midrule
16 & $1.520\;[1.236,1.812]$ & $1.621\;[1.330,1.921]$ & $0.948\;[0.731,1.175]$ & $1.379\;[1.110,1.661]$\\
20 & $0.784\;[0.627,0.937]$ & $0.983\;[0.781,1.186]$ & $0.780\;[0.605,0.962]$ & $1.156\;[0.925,1.400]$\\
22 & $0.860\;[0.665,1.059]$ & $1.015\;[0.782,1.254]$ & $0.915\;[0.716,1.125]$ & $1.083\;[0.849,1.329]$\\
23 & $0.565\;[0.396,0.735]$ & $0.670\;[0.459,0.881]$ & $0.405\;[0.272,0.542]$ & $0.511\;[0.350,0.676]$\\
24 & $0.295\;[0.188,0.403]$ & $0.351\;[0.233,0.472]$ & $0.232\;[0.126,0.337]$ & $0.301\;[0.179,0.423]$\\
\bottomrule
\end{tabularx}
\end{table}

\paragraph{Orthogonal state edits and their joint effect on the answer.}
There is a direct test with the same receiver and donor. Let $H$ be the unedited suffix at the chosen layer, $x$ and $y$ its projected question and content increments, and $M$ the final updated-minus-original margin for that receiver. Subtract the effect of the content edit alone from its additional effect in the presence of the question edit:
\[
I=\big[M(H+x+y)-M(H+x)\big]-\big[M(H+y)-M(H)\big].
\]
All four margins are observed; $x$ and $y$ denote requested edits before native rounding. Orthogonality is defined on the state vectors, not on their effects on predictions \citep{park2024linear}. Here $I$ tests the latter. An affine decoder would give zero in exact arithmetic. For a twice continuously differentiable real-valued decoder the same contrast equals
\[
I=\int_0^1\!\int_0^1 x^\top\nabla^2 M(H+s x+t y)\,y\,dt\,ds.
\]
For example, both edits could add evidence to the same intermediate score. If later computation maps that score to the logit margin with diminishing slope, the second positive increment has a smaller effect in the presence of the first. This would give $I<0$ without requiring a separate source gate. We measure four logit margins, not curvature or saturation. Native rounding also enters the measured contrast; it does not locate the interaction in a particular layer.

At layer 23, with question-first fitting, $I$ is $-0.119$ logits in MOSS (95\% CI $[-0.228,-0.014]$) and $-0.110$ in MiMo ($[-0.158,-0.064]$). Both content increments still favor the unasked answer, but less than when added alone. With content-first fitting, $I$ is $-0.186$ in MOSS ($[-0.292,-0.083]$) and $-0.029$ in MiMo ($[-0.0581,0.0006]$); the latter interval includes zero. These are pointwise intervals conditional on the fitted bases. The experiment distinguishes a reproducible direction of influence across fits from two independent semantic contributions. It does not show that the unedited model first selects a source and then reads its content.

Changing the fitting order changes which contrast supplies the coefficient on shared directions. It matters to the answer only insofar as the resulting state change reaches the distinction between the two answer labels. Across orders, adding the content edit still favors the unasked answer; the separate effects are not counts of independent semantic features.

\subsection{Frozen-Basis Paraphrase and Same-Event Controls}

Two authored paraphrases per task family retain the person's name, event, and intended answer. The bases remain fitted only on the original wording. At the five preselected layers 16, 20, 22, 23, and 24, we separately change the receiving question, change the donor question used to form its contrast, or transfer a wording difference that keeps the queried event unchanged. All conditions cover the same 512 scenarios and both basis orders. Tables~\ref{tab:subspace-paraphrase} and~\ref{tab:subspace-wording} give every question-contrast comparison at these layers.

The question-contrast effect persists under paraphrasing, but its magnitude is not uniformly invariant. At layer 23 with question-first construction, the original, p1, and p2 effects are 2.154, 2.117, and 1.969 in MOSS and 0.811, 0.726, and 0.813 in MiMo. All six intervals exclude zero. Receiving paraphrase p2 changes the MOSS effect by $-0.185$ ($[-0.243,-0.128]$); receiving paraphrase p1 changes the MiMo effect by $-0.084$ ($[-0.115,-0.054]$). Same-event wording controls also produce nonzero effects in several conditions. These controls show transfer across the tested phrasings, not that the directions encode source identity alone: paraphrases preserve answer content as well as the requested event. Section~\ref{sec:mechanism} tests information recoverability and a connected computational path without requiring that semantic decomposition.

\begin{table}[htbp]
\centering
\caption{Question and content edits at layer 23 on original-plan questions. The margin is the updated minus original answer logit; brackets give pointwise 95\% intervals over 512 audio groups. Top: differences between question-only shifts. Paraphrases p1/p2 preserve the event and answer and use the question-first basis. Bottom: content shifts with that basis. Edit magnitudes are not matched across fitting orders.}
\label{tab:contrast-audit}
\tableformat
\begin{tabularx}{\linewidth}{@{}>{\raggedright\arraybackslash}X c c@{}}
\toprule
\textbf{Question edit: difference between settings} & \textbf{MOSS} & \textbf{MiMo}\\
\midrule
Content-first minus question-first & $-0.161\;[-0.206,-0.117]$ & $-0.577\;[-0.641,-0.514]$\\
Receiving-question paraphrase p1 & $-0.037\;[-0.080,0.006]$ & $-0.084\;[-0.115,-0.054]$\\
Receiving-question paraphrase p2 & $-0.185\;[-0.243,-0.128]$ & $0.003\;[-0.028,0.034]$\\
\midrule
\textbf{Content edit, question fitted first} & \textbf{MOSS} & \textbf{MiMo}\\
Content edit alone & $0.684\;[0.471,0.903]$ & $0.515\;[0.355,0.682]$\\
Additional shift after the question edit & $0.565\;[0.396,0.735]$ & $0.405\;[0.272,0.542]$\\
\bottomrule
\end{tabularx}
\end{table}

Intervals resample complete audio-connected groups 10,000 times, preserving all paired conditions. They are pointwise and conditional on the frozen bases and donor assignments; they do not include basis-refitting uncertainty or simultaneous guarantees over scanned layers.

\begin{table}[p]
\centering
\caption{Question paraphrase controls over all 512 scenarios. Entries are changes in the question-contrast effect relative to the original template, with paired 95\% intervals. The receiving-question column changes only the receiver wording; the donor-question column changes the wording defining the donor contrast. Q-first and C-first denote the two construction orders. Native next-token logits are used.}
\label{tab:subspace-paraphrase}
\tableformat
\renewcommand{\arraystretch}{1.0}
\begin{tabularx}{\linewidth}{@{}r c l *{2}{>{\centering\arraybackslash}X}@{}}
\toprule
\textbf{Layer} & \textbf{Template} & \textbf{Order} & \textbf{Receiving question} & \textbf{Donor question}\\
\midrule
\multicolumn{5}{l}{\textbf{MOSS}}\\
16 & p1 & Q-first & $0.018\;[-0.020,0.056]$ & $-0.201\;[-0.228,-0.174]$\\
16 & p1 & C-first & $0.010\;[-0.029,0.049]$ & $-0.183\;[-0.207,-0.159]$\\
16 & p2 & Q-first & $0.080\;[0.036,0.127]$ & $-0.179\;[-0.206,-0.152]$\\
16 & p2 & C-first & $0.095\;[0.051,0.142]$ & $-0.146\;[-0.172,-0.120]$\\
20 & p1 & Q-first & $-0.195\;[-0.250,-0.141]$ & $-0.077\;[-0.115,-0.040]$\\
20 & p1 & C-first & $-0.163\;[-0.216,-0.111]$ & $-0.054\;[-0.083,-0.024]$\\
20 & p2 & Q-first & $-0.083\;[-0.140,-0.025]$ & $-0.233\;[-0.304,-0.165]$\\
20 & p2 & C-first & $-0.064\;[-0.115,-0.012]$ & $-0.169\;[-0.222,-0.117]$\\
22 & p1 & Q-first & $-0.108\;[-0.151,-0.065]$ & $-0.064\;[-0.095,-0.033]$\\
22 & p1 & C-first & $-0.092\;[-0.133,-0.050]$ & $-0.043\;[-0.071,-0.015]$\\
22 & p2 & Q-first & $-0.151\;[-0.206,-0.096]$ & $-0.198\;[-0.260,-0.138]$\\
22 & p2 & C-first & $-0.114\;[-0.165,-0.064]$ & $-0.144\;[-0.197,-0.092]$\\
23 & p1 & Q-first & $-0.037\;[-0.080,0.006]$ & $-0.041\;[-0.073,-0.008]$\\
23 & p1 & C-first & $-0.033\;[-0.073,0.007]$ & $-0.029\;[-0.058,0.001]$\\
23 & p2 & Q-first & $-0.185\;[-0.243,-0.128]$ & $-0.242\;[-0.303,-0.184]$\\
23 & p2 & C-first & $-0.136\;[-0.188,-0.085]$ & $-0.201\;[-0.256,-0.150]$\\
24 & p1 & Q-first & $0.045\;[0.004,0.085]$ & $0.030\;[-0.002,0.062]$\\
24 & p1 & C-first & $0.040\;[-0.003,0.082]$ & $0.034\;[0.003,0.065]$\\
24 & p2 & Q-first & $-0.172\;[-0.230,-0.114]$ & $-0.104\;[-0.158,-0.049]$\\
24 & p2 & C-first & $-0.175\;[-0.233,-0.118]$ & $-0.109\;[-0.162,-0.055]$\\
\midrule
\multicolumn{5}{l}{\textbf{MiMo}}\\
16 & p1 & Q-first & $-0.031\;[-0.049,-0.014]$ & $-0.020\;[-0.035,-0.005]$\\
16 & p1 & C-first & $-0.011\;[-0.028,0.005]$ & $-0.038\;[-0.051,-0.024]$\\
16 & p2 & Q-first & $0.011\;[-0.011,0.033]$ & $0.007\;[-0.006,0.021]$\\
16 & p2 & C-first & $0.017\;[-0.003,0.038]$ & $-0.003\;[-0.015,0.010]$\\
20 & p1 & Q-first & $0.016\;[-0.017,0.048]$ & $0.070\;[0.042,0.099]$\\
20 & p1 & C-first & $-0.005\;[-0.033,0.021]$ & $0.037\;[0.015,0.059]$\\
20 & p2 & Q-first & $0.031\;[-0.006,0.066]$ & $-0.005\;[-0.031,0.019]$\\
20 & p2 & C-first & $0.036\;[0.006,0.067]$ & $-0.014\;[-0.033,0.004]$\\
22 & p1 & Q-first & $0.015\;[-0.014,0.044]$ & $-0.012\;[-0.037,0.015]$\\
22 & p1 & C-first & $0.004\;[-0.024,0.033]$ & $-0.034\;[-0.060,-0.008]$\\
22 & p2 & Q-first & $0.032\;[-0.002,0.067]$ & $-0.061\;[-0.084,-0.037]$\\
22 & p2 & C-first & $0.021\;[-0.014,0.056]$ & $-0.072\;[-0.095,-0.048]$\\
23 & p1 & Q-first & $-0.084\;[-0.115,-0.054]$ & $-0.055\;[-0.073,-0.037]$\\
23 & p1 & C-first & $-0.044\;[-0.061,-0.026]$ & $-0.054\;[-0.067,-0.042]$\\
23 & p2 & Q-first & $0.003\;[-0.028,0.034]$ & $-0.037\;[-0.055,-0.019]$\\
23 & p2 & C-first & $0.004\;[-0.013,0.022]$ & $-0.064\;[-0.077,-0.052]$\\
24 & p1 & Q-first & $-0.063\;[-0.082,-0.044]$ & $-0.031\;[-0.043,-0.018]$\\
24 & p1 & C-first & $-0.047\;[-0.063,-0.030]$ & $-0.022\;[-0.034,-0.011]$\\
24 & p2 & Q-first & $0.020\;[0.000,0.040]$ & $-0.030\;[-0.043,-0.018]$\\
24 & p2 & C-first & $-0.014\;[-0.031,0.003]$ & $-0.006\;[-0.017,0.005]$\\
\bottomrule
\end{tabularx}
\end{table}

\begin{table}[p]
\centering
\caption{Same-event wording controls over all 512 scenarios. The donor increment changes wording while retaining the queried event, then projects through the question-contrast basis. Entries are effects on the updated-minus-original-event answer margin, with 95\% intervals; they are not accuracy gains. Q-first and C-first denote the two construction orders.}
\label{tab:subspace-wording}
\tableformat
\renewcommand{\arraystretch}{1.0}
\begin{tabularx}{\linewidth}{@{}r c l *{2}{>{\centering\arraybackslash}X}@{}}
\toprule
\textbf{Layer} & \textbf{Template} & \textbf{Order} & \textbf{Original-event wording} & \textbf{Updated-event wording}\\
\midrule
\multicolumn{5}{l}{\textbf{MOSS}}\\
16 & p1 & Q-first & $-0.005\;[-0.021,0.011]$ & $-0.051\;[-0.070,-0.033]$\\
16 & p1 & C-first & $-0.004\;[-0.020,0.012]$ & $-0.029\;[-0.047,-0.010]$\\
16 & p2 & Q-first & $0.023\;[0.006,0.040]$ & $-0.037\;[-0.056,-0.018]$\\
16 & p2 & C-first & $0.025\;[0.007,0.043]$ & $-0.002\;[-0.019,0.016]$\\
20 & p1 & Q-first & $0.003\;[-0.026,0.031]$ & $-0.065\;[-0.088,-0.043]$\\
20 & p1 & C-first & $0.010\;[-0.014,0.035]$ & $-0.053\;[-0.073,-0.031]$\\
20 & p2 & Q-first & $0.107\;[0.060,0.154]$ & $-0.064\;[-0.097,-0.031]$\\
20 & p2 & C-first & $0.117\;[0.076,0.159]$ & $-0.038\;[-0.067,-0.009]$\\
22 & p1 & Q-first & $0.009\;[-0.014,0.032]$ & $-0.031\;[-0.055,-0.006]$\\
22 & p1 & C-first & $0.014\;[-0.005,0.034]$ & $0.008\;[-0.014,0.031]$\\
22 & p2 & Q-first & $0.106\;[0.068,0.145]$ & $0.006\;[-0.027,0.040]$\\
22 & p2 & C-first & $0.121\;[0.088,0.155]$ & $0.064\;[0.034,0.094]$\\
23 & p1 & Q-first & $-0.002\;[-0.023,0.019]$ & $-0.039\;[-0.062,-0.016]$\\
23 & p1 & C-first & $0.007\;[-0.012,0.026]$ & $-0.035\;[-0.057,-0.013]$\\
23 & p2 & Q-first & $0.114\;[0.079,0.150]$ & $-0.039\;[-0.071,-0.007]$\\
23 & p2 & C-first & $0.126\;[0.094,0.160]$ & $-0.010\;[-0.041,0.021]$\\
24 & p1 & Q-first & $-0.024\;[-0.045,-0.003]$ & $-0.016\;[-0.045,0.012]$\\
24 & p1 & C-first & $-0.019\;[-0.040,0.001]$ & $-0.014\;[-0.041,0.013]$\\
24 & p2 & Q-first & $0.029\;[-0.011,0.068]$ & $-0.039\;[-0.095,0.014]$\\
24 & p2 & C-first & $0.028\;[-0.011,0.066]$ & $-0.045\;[-0.099,0.006]$\\
\midrule
\multicolumn{5}{l}{\textbf{MiMo}}\\
16 & p1 & Q-first & $0.005\;[-0.006,0.016]$ & $0.010\;[-0.001,0.020]$\\
16 & p1 & C-first & $0.016\;[0.006,0.027]$ & $0.008\;[-0.000,0.017]$\\
16 & p2 & Q-first & $-0.025\;[-0.036,-0.015]$ & $0.016\;[0.006,0.025]$\\
16 & p2 & C-first & $-0.015\;[-0.025,-0.005]$ & $0.012\;[0.004,0.021]$\\
20 & p1 & Q-first & $0.029\;[0.014,0.045]$ & $0.006\;[-0.012,0.025]$\\
20 & p1 & C-first & $0.038\;[0.023,0.052]$ & $0.021\;[0.004,0.039]$\\
20 & p2 & Q-first & $-0.009\;[-0.024,0.006]$ & $0.015\;[0.001,0.029]$\\
20 & p2 & C-first & $-0.002\;[-0.016,0.013]$ & $0.025\;[0.012,0.038]$\\
22 & p1 & Q-first & $0.033\;[0.017,0.048]$ & $-0.048\;[-0.065,-0.030]$\\
22 & p1 & C-first & $0.025\;[0.011,0.039]$ & $-0.051\;[-0.067,-0.035]$\\
22 & p2 & Q-first & $0.006\;[-0.011,0.023]$ & $-0.045\;[-0.060,-0.029]$\\
22 & p2 & C-first & $0.006\;[-0.009,0.022]$ & $-0.050\;[-0.064,-0.036]$\\
23 & p1 & Q-first & $-0.020\;[-0.032,-0.008]$ & $-0.058\;[-0.073,-0.044]$\\
23 & p1 & C-first & $-0.019\;[-0.029,-0.010]$ & $-0.065\;[-0.079,-0.051]$\\
23 & p2 & Q-first & $-0.030\;[-0.042,-0.017]$ & $-0.046\;[-0.060,-0.033]$\\
23 & p2 & C-first & $-0.002\;[-0.013,0.009]$ & $-0.058\;[-0.072,-0.045]$\\
24 & p1 & Q-first & $-0.020\;[-0.029,-0.010]$ & $-0.040\;[-0.051,-0.028]$\\
24 & p1 & C-first & $-0.015\;[-0.024,-0.006]$ & $-0.034\;[-0.044,-0.023]$\\
24 & p2 & Q-first & $-0.016\;[-0.026,-0.006]$ & $-0.036\;[-0.047,-0.025]$\\
24 & p2 & C-first & $-0.003\;[-0.012,0.006]$ & $-0.021\;[-0.031,-0.011]$\\
\bottomrule
\end{tabularx}
\end{table}

\FloatBarrier

Transfer to a paraphrase shows that the sign of the question effect does not require the original sentence verbatim at the illustrated layer under question-first fitting. The same-event control also moves scores in some conditions. The transferable effect therefore cannot be identified with a source label alone.

\subsection{Paraphrase Templates}

Table~\ref{tab:paraphrase-templates} lists the two authored paraphrases used for each task family. The placeholder \emph{name} is replaced by the same scenario name in every condition. The original template, rather than either paraphrase, supplies the fitting inputs. These templates change question wording, not the past utterances, person, or requested original/updated arrangement.

\begin{table}[!htbp]
\centering
\caption{Exact paraphrase templates. Each row pairs questions about the original and updated arrangements. These are the wording controls, not source-index instructions.}
\label{tab:paraphrase-templates}
\tableformat
\begin{tabularx}{\linewidth}{@{}l c *{2}{>{\raggedright\arraybackslash}X}@{}}
\toprule
\textbf{Family} & \textbf{ID} & \textbf{Original arrangement} & \textbf{Updated arrangement}\\
\midrule
Time & p1 & What was \emph{name}'s originally planned departure time? & What is \emph{name}'s updated departure time?\\
& p2 & According to the original plan, when was \emph{name} going to leave? & According to the updated plan, when will \emph{name} leave?\\
\midrule
Color & p1 & What was the color of the suitcase \emph{name} initially planned to bring? & What is the color of the suitcase \emph{name} now plans to bring?\\
& p2 & According to the original plan, which suitcase color did \emph{name} choose? & According to the revised plan, which suitcase color did \emph{name} choose?\\
\midrule
Quantity & p1 & How many notebooks were in \emph{name}'s initial purchase plan? & How many notebooks are in \emph{name}'s updated purchase plan?\\
& p2 & According to the original plan, how many notebooks was \emph{name} going to buy? & According to the updated plan, how many notebooks will \emph{name} buy?\\
\midrule
Location & p1 & Where was \emph{name}'s passport placed initially? & Where is \emph{name}'s passport located after it was moved?\\
& p2 & What was the initial location of \emph{name}'s passport? & What is the current location of \emph{name}'s passport?\\
\bottomrule
\end{tabularx}
\end{table}

\FloatBarrier

The wording pairs keep the event and its answer fixed when testing paraphrases, and change the requested arrangement when testing question contrasts. This separates sensitivity to rewording from sensitivity to changing the question's referent; the latter contrast still changes which answer is correct.

The stable result is conditional: the content of a past utterance can add support for the unasked answer after a question edit has been applied. Its measured size depends on which directions were assigned first and on how the decoder reads them. A transferable effect is evidence of influence, not a unique decomposition into a source signal and an answer signal.

\end{document}